%% file: main.tex
\documentclass[10pt,twocolumn,letterpaper]{article}

\usepackage[pagenumbers]{cvpr} 

\usepackage{colortbl}
\usepackage{xcolor}

\input{preamble}

\definecolor{cvprblue}{rgb}{0.21,0.49,0.74}
\usepackage[pagebackref,breaklinks,colorlinks,allcolors=cvprblue]{hyperref}

\def\paperID{} 
\def\confName{CVPR}
\def\confYear{2026}

\title{Beyond Static Frames: Temporal Aggregate-and-Restore Vision Transformer for Human Pose Estimation}

\author{
Hongwei Fang, Jiahang Cai, Xun Wang, Wenwu Yang\footnotemark[1]\\
Zhejiang Gongshang University, China
}

\begin{document}
\maketitle
\footnotetext[1]{Corresponding author (wwyang@zjgsu.edu.cn).}
\input{new_sec/0_abstract}

\input{new_sec/1_intro}

\input{new_sec/2_relatedwork}

\input{new_sec/3_method}
\input{new_sec/4_result}

\section{Conclusion}
In this work, we introduced TAR-ViTPose, a temporal aggregate-and-restore framework that extends plain ViTs to video-based 2D HPE. By integrating joint-centric temporal aggregation (JTA) for aligning keypoint features across frames and global restoring attention (GRA) for reinjecting temporal cues into the current-frame representation, our method effectively leverages temporal coherence while preserving the simplicity and efficiency of the ViT architecture. Extensive experiments show that TAR-ViTPose substantially improves over the single-frame ViTPose baseline and surpasses existing SOTA video-based methods, while also delivering superior runtime performance. These results demonstrate the strong potential of plain ViTs for efficient and accurate real-world video pose estimation.

\section{Acknowledgments}
We would like to thank Runyang Feng for the helpful discussions.
This work was supported by NSF of China (62541209) and the Fundamental Research Funds for the Provincial Universities of Zhejiang (FR24005Z). 

{
    \small
    \bibliographystyle{ieeenat_fullname}
    \bibliography{main}
}

\input{new_sec/X_suppl}

\end{document}

%% file: preamble.tex
\usepackage{multirow}
\usepackage{graphicx}
\usepackage{makecell}
\usepackage{pifont}
\usepackage{amsmath}

%% file: new_sec/0_abstract.tex
\begin{abstract}

Vision Transformers (ViTs) have recently achieved state-of-the-art performance in 2D human pose estimation due to their strong global modeling capability. However, existing ViT-based pose estimators are designed for static images and process each frame independently, thereby ignoring the temporal coherence that exists in video sequences. This limitation often results in unstable predictions, especially in challenging scenes involving motion blur, occlusion, or defocus. In this paper, we propose \mbox{TAR-ViTPose}, a novel Temporal Aggregate-and-Restore Vision Transformer tailored for video-based 2D human pose estimation. \mbox{TAR-ViTPose} enhances static ViT representations by aggregating temporal cues across frames in a plug-and-play manner, leading to more robust and accurate pose estimation. To effectively aggregate joint-specific features that are temporally aligned across frames, we introduce a joint-centric temporal aggregation (JTA) that assigns each joint a learnable query token to selectively attend to its corresponding regions from neighboring frames. Furthermore, we develop a global restoring attention (GRA) to restore the aggregated temporal features back into the token sequence of the current frame, enriching its pose representation while fully preserving global context for precise keypoint localization. Extensive experiments demonstrate that \mbox{TAR-ViTPose} substantially improves upon the single-frame baseline \mbox{ViTPose}, achieving a +2.3 mAP gain on the PoseTrack2017 benchmark. Moreover, our approach outperforms existing state-of-the-art video-based methods, while also achieving a noticeably higher runtime frame rate in real-world applications. Project page: \textcolor{magenta}{https://github.com/zgspose/TARViTPose}.
\end{abstract}

%% file: new_sec/1_intro.tex
\section{Introduction}
\label{sec:intro}

Human pose estimation (HPE) aims to localize anatomical keypoints of individuals in images or videos. As a core task in computer vision, HPE plays a vital role in numerous human-centric applications, including human-computer interaction~\cite{HOI-CVPR2020,chen2025prvr}, human behavior analysis~\cite{dong2023hierarchical}, and motion capture~\cite{MOCap-CVPR2020}. 
Although HPE is primarily deployed in video-based scenarios, the majority of existing methods rely on single-frame approaches~\cite{TransPose_ICCV2021,TokenPose_CVPR2023,CID_cvpr2022,Vitpose_NIPS2022,HRNet_CVPR2019,SimplePose_ECCV2018,RLE_ICCV2021,PEDR-CVPR2022,PoseLLM-CVPR2024,Sapiens-ECCV2024} that perform 2D pose estimation independently on\textit{ static images}. However, these approaches inherently lack temporal awareness and are thus susceptible to failure in dynamic scenarios.
Recently, several video-based HPE methods~\cite{PoseWarper_NIPS2019,DCPose_CVPR2021,OTPose_SMC2022,TDMI_CVPR2023,DSTA-CVPR2024,CMPose-AAAI2025,HRSM-ICCV2025} have been developed to address this limitation by exploiting \textit{dynamic cues} across video frames.
These studies demonstrate that incorporating temporal information is crucial for improving robustness under challenging conditions such as occlusion, motion blur, and video defocus.
\begin{figure}
    \centering
    \includegraphics[width=1.0\linewidth]{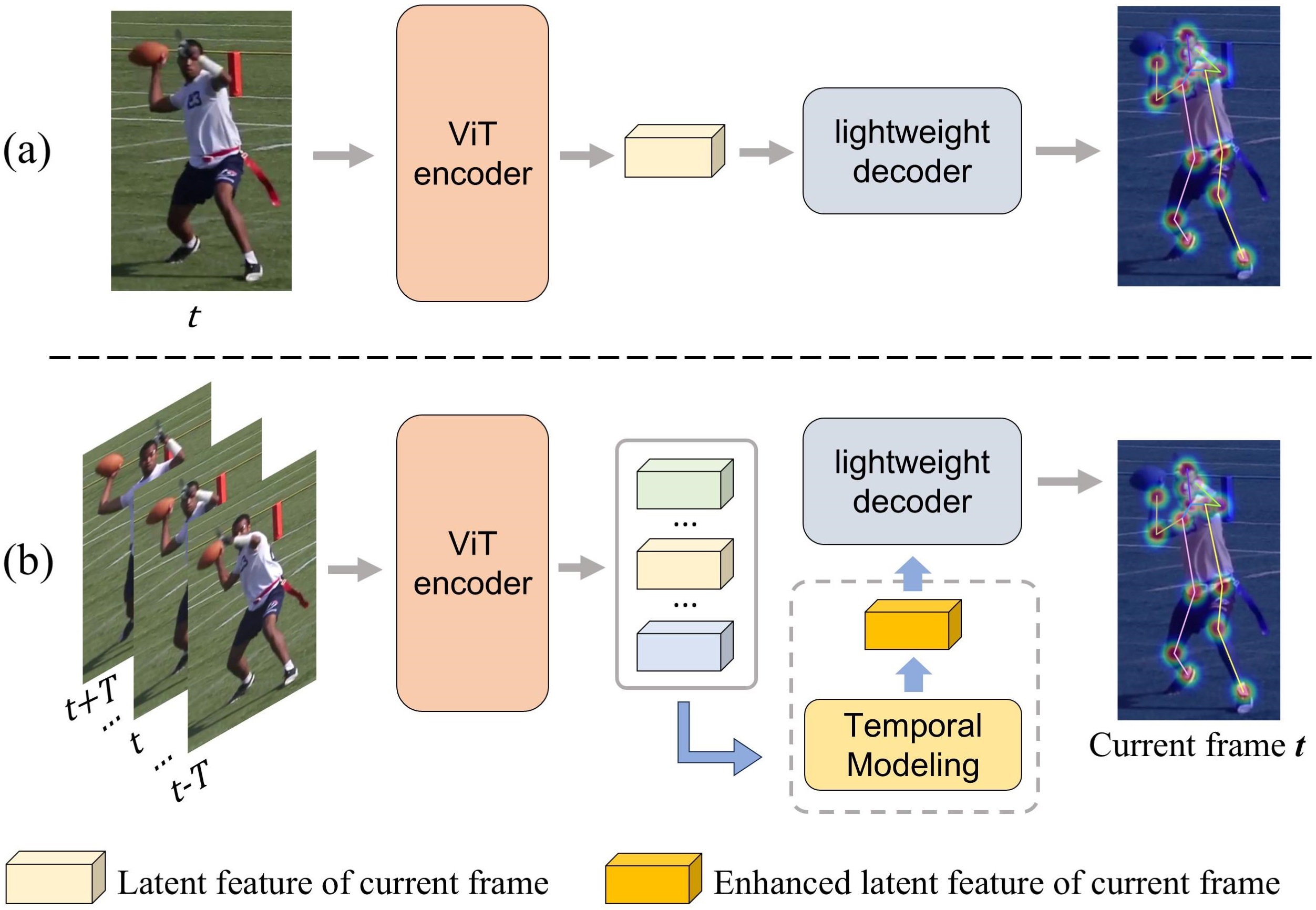}
    \caption{Comparison between the baseline ViTPose pipeline (a) and our TAR-ViTPose (b). (a) ViTPose adopts a ViT encoder to extract latent features from the input image, which are then fed into a lightweight decoder to regress keypoint heatmaps. (b) Our method enhances the current-frame representation by aggregating temporal cues from adjacent frames, achieving plug-and-play temporal modeling within the original ViTPose architecture.}
    \label{fig:demo}
    \vspace{-4mm}
\end{figure}

Existing video-based methods are mainly designed with convolutional neural network architectures. Recently, Vision Transformers (ViTs) have been introduced to significantly improve single-frame pose estimation performance. Among them, ViTPose~\cite{Vitpose_NIPS2022}, built upon a plain ViT encoder and a lightweight decoder, has emerged as a state-of-the-art 2D pose estimator, exhibiting strong spatial modeling capability and remarkable generalization. Inspired by the strong performance of ViTPose, several recent video-based methods have attempted to reuse it for temporal pose estimation. However, these approaches typically employ the pre-trained ViTPose solely for single-frame feature extraction, while relying on additional, carefully designed modules—such as Transformer-based architectures (\textit{e.g.}, DSTA~\cite{DSTA-CVPR2024}, {CM-Pose}~\cite{CMPose-AAAI2025}, MTPose~\cite{MTPose-ijcai2024}) or Mamba-based models (\textit{e.g.}, GLSMamba~\cite{HRSM-ICCV2025})—to fuse multi-frame features, along with a dedicated decoder to adapt to the fused representations. This not only complicates the pipeline and increases inference cost but also deviates from the simplicity of the plain ViT architectures. This motivates us to explore a new direction: \textit{can we directly embed temporal modeling into the \mbox{ViTPose} framework, while preserving its 
plain ViT design and lightweight decoding pipeline?}

To answer this question, we propose \mbox{TAR-ViTPose}, a novel Temporal Aggregate-and-Restore Vision Transformer that enhances static ViT representations by aggregating temporal cues across frames in a plug-and-play manner, thereby enabling more robust and accurate pose estimation. As illustrated in Fig.~\ref{fig:demo}, our strategy is to append a temporal modeling module after the \mbox{ViTPose} encoder to aggregate latent features from adjacent frames into the current frame's feature tokens, and then reuse the original lightweight decoder of \mbox{ViTPose} to regress keypoint heatmaps from the enhanced representation. 
The key challenge lies in ensuring that temporally corresponding keypoint features are accurately aligned across frames within the temporal modeling module, so as to effectively capture joint-specific temporal dynamics. To address this, we introduce a Joint-centric Temporal Aggregation (JTA) that assigns each joint a learnable query token and employs a mask-aware attention mechanism to selectively attend to its corresponding regions in neighboring frames. In addition, we develop a Global Restoring Attention (GRA) to restore the aggregated temporal features back into the token sequence of the current frame, enriching its pose representation while fully preserving global context for precise keypoint localization.

We extensively evaluate our method on three widely used video-based HPE benchmarks: PoseTrack2017~\cite{PoseTrack2017_CVPR2017}, PoseTrack2018~\cite{PoseTrack2018_CVPR2018}, and PoseTrack21~\cite{PoseTrack21_CVPR2022}. Experimental results demonstrate that our \mbox{TAR-ViTPose} framework achieves substantial performance gains, surpassing the single-frame baseline \mbox{ViTPose} by a notable margin of \textbf{+2.3} mAP on PoseTrack2017 benchmark. Moreover, our approach sets new state-of-the-art results on all three benchmarks, while achieving higher runtime frame rates \textit{(e.g.}, \textbf{413} fps \textit{vs.} 52 fps ) in real-world applications compared to existing video-based methods.
It is worth noting that our work does not address temporal pose tracking; instead, it presents a simple and robust temporal Vision Transformer framework with superior performance for 2D pose estimation in videos.

Our main contributions can be summarized as follows:
\begin{itemize}
\item We propose \textbf{TAR-ViTPose}, a novel Temporal Aggregate-and-Restore Vision Transformer that integrates temporal modeling into the ViTPose framework in a plug-and-play fashion, while preserving its plain ViT design and lightweight decoding pipeline. By leveraging temporal cues across frames, our method significantly enhances pose estimation robustness and accuracy.
\item We introduce Joint-centric Temporal Aggregation (JTA) and Global Restoring Attention (GRA), which together enable effective aggregation and reintegration of joint-specific temporal features across frames.
\item Extensive experiments on video-based benchmarks show that our method achieves substantial performance gains over the single-frame baseline ViTPose, establishes new state-of-the-art results, and delivers higher runtime frame rates than existing video-based methods.
\end{itemize}

%% file: new_sec/2_relatedwork.tex
\section{Related Work}
\label{sec:related_work}

\subsection{Image-Based Human Pose Estimation} 
Early approaches to human pose estimation primarily relied on deformable models and pictorial structures to capture spatial relationships among body parts~\cite{FirstHeatmap_NIPS2014, CPS_ECCV2010}.
With the advent of deep learning, convolutional neural networks (CNNs) have significantly advanced this field~\cite{CPM-CVPR2016,SHN-ECCV2016,OpenPose-CVPR2017,SimplePose_ECCV2018,HRNet_CVPR2019,CID_cvpr2022,RLE_ICCV2021,PEDR-CVPR2022}.
More recently, the emergence of the Vision Transformer (ViT) has further reshaped the landscape of human pose estimation from static images.
Methods like TransPose~\cite{TransPose_ICCV2021}, TokenPose~\cite{TokenPose_CVPR2023}, HRFormer~\cite{HRFormer_NIPS2021}, and the highly influential ViTPose~\cite{Vitpose_NIPS2022} employ global self-attention mechanisms to capture long-range dependencies among human joints, setting new benchmarks across multiple 2D HPE datasets.
Among these ViT-based approaches, TransPose~\cite{TransPose_ICCV2021}, TokenPose~\cite{TokenPose_CVPR2023}, and HRFormer~\cite{HRFormer_NIPS2021} either rely on additional CNN backbones for feature extraction or require carefully designed Transformer structures to adapt to the pose estimation task.
In contrast, ViTPose~\cite{Vitpose_NIPS2022} adopts the plain and non-hierarchical ViT architecture without task-specific structural modifications, thereby establishing a simple yet highly effective ViT-based baseline model for human pose estimation.
However, despite their outstanding performance on static images, these methods inherently lack temporal modeling capabilities, which limits their applicability in dynamic video scenarios.
Consequently, such models struggle to handle challenges unique to video inputs, such as motion blur, defocus, and pose occlusions, which frequently occur in real-world  scenes.

\begin{figure*}
    \centering
    \includegraphics[width=1.0\linewidth]{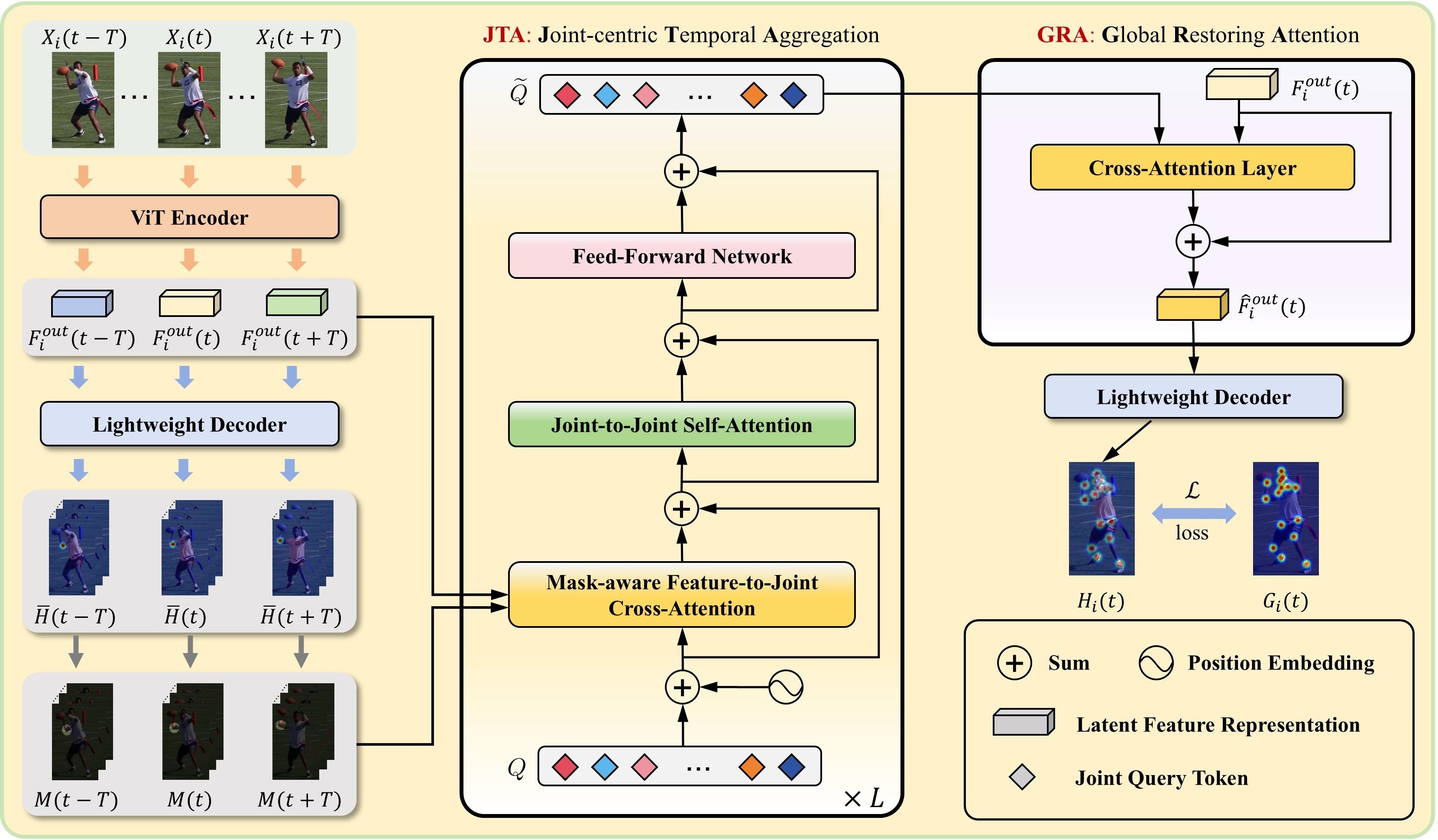}
    \caption{The pipeline of the proposed Temporal Aggregate-and-Restore Vision Transformer (TAR-ViTPose). 
        The objective is to estimate the human pose of the current frame \( X_i(t) \). 
        Given a video sequence 
        \(\langle X_i(t\!-\!T), \dots, X_i(t), \dots, X_i(t\!+\!T) \rangle\), 
        each frame is first encoded by the ViT encoder to obtain latent features 
        \(\{ F_i^{\text{out}}(\tau) \}_{\tau = t-T}^{t+T}\). 
        The JTA  precisely aligns and aggregates keypoint features across frames. 
        To achieve this, a query token is assigned to each joint ($Q$), and a mask-aware attention  selectively attends to its corresponding joint regions in neighboring frames, guided by masks 
        \( M(\tau)_{\tau = t-T}^{t+T} \) 
        derived from the decoded keypoint heatmaps \( \overline{H}(\tau)_{\tau = t-T}^{t+T} \). 
        Subsequently, the GRA injects the aggregated temporal features \(\widetilde{Q}\) back into the current frame's latent representation, producing an enhanced feature \(\widehat{F}_i^{\text{out}}(t)\), 
        which is then fed into the lightweight decoder to generate the final keypoint heatmaps $H_i(t)$ for the current frame.
    }
    \label{fig:pipeline}
    \vspace{-5mm}
\end{figure*}

\subsection{Video-Based Human Pose Estimation} 
Compared to methods designed for static images, video-based 2D HPE approaches~\cite{PoseFlow-BMVC2018,PoseWarper_NIPS2019,DCPose_CVPR2021,OTPose_SMC2022,TDMI_CVPR2023,DSTA-CVPR2024,CMPose-AAAI2025,HRSM-ICCV2025,Poseidon-Arxiv2025} are more suitable for motion-rich and dynamic environments.
These video-based methods are mainly designed with 
convolutional neural network architectures. 
Inspired by the strong performance of the ViT-based 2D HPE method \mbox{ViTPose}, several recent video-based methods have attempted to extend it for temporal pose estimation.
However, these approaches typically employ the pre-trained ViTPose solely for single-frame feature extraction, while relying on additional, carefully designed modules such as Transformer-based architectures (\textit{e.g.}, DSTA~\cite{DSTA-CVPR2024}, CM-Pose~\cite{CMPose-AAAI2025}, MTPose~\cite{MTPose-ijcai2024}) or Mamba-based models (\textit{e.g.}, GLSMamba~\cite{HRSM-ICCV2025}) to fuse multi-frame features, along with dedicated decoders to adapt to the fused representations.
This not only complicates the overall pipeline and increases inference cost but also deviates from the simplicity and elegance of the plain ViT architecture employed in ViTPose.
Although Poseidon~\cite{Poseidon-Arxiv2025} reuses the lightweight decoder of ViTPose, it merely fuses multi-frame features through simple cross attention, which struggles to accurately align temporally corresponding keypoint features across frames.

To date, there have been few efforts to explore the full potential of plain ViTs for spatiotemporal pose estimation.
In this work, we address this gap by proposing a simple yet effective baseline that directly integrates temporal modeling into the ViTPose framework, while preserving its plain ViT design and lightweight decoding pipeline.

%% file: new_sec/3_method.tex
\section{Our Approach}

Let $X(t)$ denote the video frame at the current time 
\textit{t}, which contains multiple individuals. Our goal is to estimate the 2D locations of anatomical keypoints for each person in this frame.
Unlike single-frame approaches such as ViTPose~\cite{Vitpose_NIPS2022}, which perform pose estimation independently on static images, we leverage temporal dynamics from a sequence of consecutive frames $\mathbf{S} = \langle X(t-T), \dots, X(t), \dots, X(t+T) \rangle$, 
where $T$ denotes a predefined temporal span, to enhance pose estimation for the current frame $X(t)$.
Our method follows the widely adopted two-stage top-down paradigm. In the first stage, we apply a human detector~\cite{FastRCNN-NIPS2015} to localize individuals in the current video frame $X(t)$. Each detected bounding box is then expanded by $25\%$ to crop the corresponding person region across all frames in the sequence $\mathbf{S}$, resulting in a person-specific video clip $\mathbf{S}_i = \langle X_i(t-T), \dots, X_i(t), \dots, X_i(t+T) \rangle$ for each individual $i$. In the second stage, human pose estimation is performed for each individual $i$ within the current frame $X(t)$, which can be expressed as:
\begin{equation}
\left\{ H_i^j(t) \right\}_{j=1}^N = \text{HPE}(\mathbf{S}_i),
\end{equation}
where $\text{HPE}(\cdot)$ denotes the human pose estimation module, $H_i^j(t)$ is the heatmap for the $j$-th keypoint of individual $i$ in the current frame \textit{t}, and $N$ is the number of keypoints (\textit{e.g.}, \textit{N} = 15 in the PoseTrack datasets~\cite{PoseTrack2017_CVPR2017,PoseTrack2018_CVPR2018,PoseTrack21_CVPR2022}).
For simplicity in the following description of our algorithm, unless
 otherwise specified, we will refer to a specific individual $i$.

In our implementation, the human pose estimation module $\text{HPE}(\cdot)$ adopts a ViT-based framework similar to \mbox{ViTPose}~\cite{Vitpose_NIPS2022}, retaining its plain ViT design and lightweight decoding pipeline. Meanwhile, it seamlessly integrates temporal modeling to fully exploit joint-specific temporal dynamics within the video clip $\mathbf{S}_i$, thereby improving pose estimation robustness and accuracy, especially in challenging scenarios involving motion blur, occlusion, or defocus.
 

\subsection{ViTPose Revisited and Beyond}
ViTPose~\cite{Vitpose_NIPS2022} is a state-of-the-art 2D human pose estimation framework built upon a plain Vision Transformer architecture. It replaces traditional CNN backbones such as ResNet~\cite{ResNet_CVPR2016} or HRNet~\cite{HRNet_CVPR2019} with a ViT \textit{encoder}~\cite{ViT_ICLR2021}, in which the input image is divided into fixed-size patches (\textit{e.g.}, $16 \times 16$), linearly projected into tokens, and then processed by standard ViT layers. The \textit{decoder} is lightweight, consisting of two stacked deconvolution layers that progressively upsample the ViT feature maps, followed by a $1 \times 1$ convolution to generate pixel-wise keypoint heatmaps.

While ViTPose excels in static image scenarios, it overlooks the temporal coherence inherent in video sequences.
When directly applied to videos, it tends to produce suboptimal predictions due to its inability to capture temporal dependencies across consecutive frames.
To address this temporal limitation in \mbox{ViTPose}, we develop \mbox{TAR-ViTPose} by introducing temporal modeling into the ViT-based architecture, while preserving its original design simplicity and decoding pipeline. As shown in Fig.~\ref{fig:demo}, given a person-specific video clip $\mathbf{S}_i$, our strategy is to append a temporal modeling module after the ViT encoder to aggregate latent features from adjacent frames, thereby enhancing the current frame's feature representation before passing it to the lightweight decoder for regressing the keypoint heatmaps $\{ H_i^j(t) \}_{j=1}^n$ of the current frame $t$. 
The overall architecture and workflow are illustrated in Fig.~\ref{fig:pipeline}.
Specifically, we introduce a Joint-centric Temporal Aggregation (JTA) to effectively aggregate joint-specific features that are temporally aligned across frames. We then employ a Global Restoring Attention (GRA) to inject the aggregated temporal features back into the token sequence of the current frame, enriching its feature representation while fully preserving global context for precise keypoint localization. The following sections provide detailed descriptions of the JTA and GRA.


\subsection{Joint-centric Temporal Aggregation}
For each image \( X_i(\tau) \in \mathbb{R}^{H \times W \times 3} \) in the input person-specific video clip $\mathbf{S}_i$, where $\tau \in \{ t - T, \dots, t, \dots, t + T \}$, the ViT encoder is employed to extract its latent feature representation. Specifically, the image is first partitioned into non-overlapping patches and embedded into tokens through a patch embedding layer, yielding
\(
    F_i(\tau) \in \mathbb{R}^{\frac{H}{d} \times \frac{W}{d} \times C},
\)
where \(d\) (\textit{e.g.}, 16 by default) denotes the patch downsampling ratio, and \(C\) is the channel dimension of the token embedding. These tokens are then processed by multiple transformer layers, producing the final feature representation output by the ViT backbone: 
\(
    F_i^{\text{out}}(\tau) \in \mathbb{R}^{\frac{H}{d} \times \frac{W}{d} \times C}.
\)

The goal of JTA is to aggregate the latent features from consecutive frames, \textit{i.e.},
\(
    \{ F_i^{\text{out}}(\tau) \}_{\tau = t-T}^{t+T},
\) through temporal modeling, 
in order to enhance the feature representation \( F_i^{\text{out}}(t) \) of the current frame \(t\). Since self-attention mechanisms~\cite{SelfAttention-NIPS2017} can naturally capture temporal dependencies, a potential approach for aggregating these latent features, as explored in~\cite{Poseidon-Arxiv2025}, is to perform self-attention directly over the feature sequence
\(
    \langle F_i^{\text{out}}(t-T), \dots, F_i^{\text{out}}(t), \dots, F_i^{\text{out}}(t+T) \rangle,
\)
or to conduct cross-attention between the current frame's feature tokens \( F_i^{\text{out}}(t) \) (as queries) and those from adjacent frames (as keys and values). However, we observe that different human joints often exhibit relatively independent temporal trajectories during motion. For example, when running, the wrist joints swing back and forth, while the head joint remains mostly forward-facing. Hence, it is crucial to model temporal dependencies in a joint-specific manner to ensure that temporally corresponding keypoint features are accurately aligned across frames.

To this end, our strategy is to assign a learnable query token to each joint and selectively attend to the corresponding feature regions across consecutive frames, rather than treating all feature tokens equally, as shown in Fig.~\ref{fig:pipeline}. This design explicitly enforces joint-level temporal alignment, enabling accurate aggregation of temporally coherent features for each keypoint. To achieve this, JTA performs cross-attention between the \( N \) learnable joint query tokens \( Q \in \mathbb{R}^{N \times C} \) (where \( N \) is the number of keypoints) and the feature tokens from all frames, thereby aggregating temporal information in a joint-specific manner:
\begin{equation}
\label{eqn:jta}
    \widetilde{Q} = \text{JTA}\left(Q,  \left\{ F_i^{\text{out}}(\tau) \right\}_{\tau = t-T}^{t+T} \right),
\end{equation}
where \( \text{JTA}(\cdot,\cdot)\) denotes the joint-centric temporal aggregation,  and \( \widetilde{Q} \) represents the updated query tokens, each aggregating features corresponding to the same joint across frames. In our implementation, the \( \text{JTA}(\cdot,\cdot) \) follows the conventional transformer architecture~\cite{SelfAttention-NIPS2017}. We stack six identical cross-attention layers sequentially. In each layer, the joint query tokens first perform cross-attention over the feature sequence \( \{ F_i^{\text{out}}(\tau) \}_{\tau = t-T}^{t+T} \) (\textit{i.e.}, feature-to-joint attention), followed by a self-attention that enables interactions among the joint queries themselves (\textit{i.e.}, joint-to-joint attention). Furthermore, to ensure that each joint query token attends only to its corresponding keypoint regions across frames during feature-to-joint attention, we introduce a mask-aware attention mechanism detailed below.

\subsubsection{Mask-aware Feature-to-Joint Attention}

To enforce joint-level temporal alignment and aggregation, each query token in 
$Q$ is designed to attend exclusively to its corresponding keypoint regions across frames, avoiding interference from unrelated areas. 
To achieve this, we generate a spatial mask map for each joint query token on every frame, as shown in Fig.~\ref{fig:pipeline}. Specifically, we first feed the frame-wise features extracted by the ViT encoder (\textit{i.e.}, $F_i^{\text{out}}(\tau)$ with $\tau \in \{ t - T, \dots, t, \dots, t + T \}$) into the lightweight decoder to obtain the keypoint heatmaps for all frames, denoted as $\overline{H}(\tau)$, where $\tau \in \{ t - T, \dots, t, \dots, t + T \}$. These heatmaps are then used to construct joint-specific binary masks, which indicate the spatial regions each joint query token should attend to across frames:
\begin{equation}
\label{eqn:mask}
    M_{x,y}^j(\tau)=
    \begin{cases}
        0       & \text{if } \overline{H}_{x,y}^j(\tau) \geq \phi, \\
        -\infty & \text{otherwise},
    \end{cases}
\end{equation}
where $\overline{H}_{x,y}^j(\tau)$ denotes the value of the $j$-th keypoint heatmap at pixel location $(x, y)$ in frame $\tau$. The resulting mask $M_{x,y}^j(\tau)$ specifies whether the $j$-th joint query token should attend to the feature at position $(x, y)$ in frame~$\tau$: $0$ indicates that the region is relevant, while $-\infty$ suppresses attention to irrelevant areas. The threshold 
$\phi$ is set to 0.2 by default. Each spatial mask is then resized to match the resolution of the ViT encoder's output feature map.

Guided by these masks, during the feature-to-joint attention process, the joint query tokens 
$Q$ perform cross-attention over the feature sequence 
\( \{ F_i^{\text{out}}(\tau) \}_{\tau = t-T}^{t+T} \)
 using masked attention~\cite{MaskAttention-CVPR2022} (with a residual connection), replacing standard cross-attention:
\begin{equation}
\label{eqn:mask_attention}
    \begin{aligned}
        {Q}^{l}= 
        \text{softmax}
        \left(
            f_q(Q^{l-1})f_k(F_i^{\text{out}}) ^\top + M
        \right)f_v(F_i^{\text{out}}) + Q^{l-1}, 
    \end{aligned}
\end{equation}
where \( l \) denotes the layer index; \( Q^l \in \mathbb{R}^{N \times C} \) represents the \( N \) joint query tokens at the \( l \)-th layer, each of dimension \( C \); \( F_i^{\text{out}} \in \mathbb{R}^{\frac{H}{d} \frac{W}{d} (2T+1) \times C} \) corresponds to the concatenated feature tokens from all frames in the sequence; and \( M \in \mathbb{R}^{N \times \frac{H}{d} \frac{W}{d}(2T+1)} \) stores the joint-specific spatial masks computed using Eq.~\ref{eqn:mask}. 
Here, \( f_q(\cdot) \), \( f_k(\cdot) \), and \( f_v(\cdot) \) denote the linear transformations applied to produce the query, key, and value embeddings, respectively. Starting from the initialization  \( Q^0 = Q \), the joint query tokens are iteratively updated across layers.
The final output of the 
$\text{JTA}(\cdot,\cdot)$, defined in Eq.~\ref{eqn:jta}, is thus given by \( \widetilde{Q} = Q^6 \).

As a result, in Eq.~\ref{eqn:mask_attention}, the mask matrix 
$M$ modulates the attention scores based on the spatial masks derived in Eq.~\ref{eqn:mask}, enforcing that each query token in 
$Q$ focuses solely on its associated keypoint regions across all frames.

\subsection{Global Restoring Attention}

Through the JTA module, the updated joint query tokens \( \widetilde{Q} \) aggregate temporal contextual cues of human keypoints across frames. To leverage these temporal cues for enhancing the current frame's feature representation \( F_i^{\text{out}}(t) \), it is crucial to reintegrate the temporal semantics into the spatial features of the current frame. This leads to a spatio-temporally enhanced feature representation, denoted as \( \widehat{F}_i^{\text{out}}(t) \). 
To achieve this, our GRA performs a single cross-attention operation between the current frame's feature tokens \( F_i^{\text{out}}(t) \) and the updated joint query tokens \( \widetilde{Q} \), thereby injecting the aggregated temporal information back into the feature space of the current frame \textit{t}:
\begin{equation}
    \widehat{F}_i^{\text{out}}(t) = \text{CrossAttn}\left(F_i^{\text{out}}(t),\, \widetilde{Q},\, \widetilde{Q}\right),
\end{equation}
where \( \text{CrossAttn}(\cdot, \cdot, \cdot) \) refers to the standard cross-attention operation, with \( F_i^{\text{out}}(t) \) serving as the query and \( \widetilde{Q} \) supplying both the keys and values.

\vspace{2mm}
\noindent\textbf{Loss Function.} The spatio-temporally enhanced feature representation \( \widehat{F}_i^{\text{out}}(t) \) is then fed into the original lightweight decoder of ViTPose~\cite{Vitpose_NIPS2022} to generate the final keypoint heatmaps for the current frame, \textit{i.e.}, \( \{ H_i^j(t) \}_{j=1}^N \). During training, the entire model (ViT encoder, JTA, GRA, and the decoder) is optimized end-to-end by minimizing the discrepancy between the predicted heatmaps and the ground-truth keypoint heatmaps:
\begin{equation}
\mathcal{L} = \sum_{i} \sum_{j=1}^{N} \left| H_i^j(t) - G_i^j(t) \right|_2^2,
\label{eqn:loss}
\end{equation}
where $ G_i^j(t)$ denotes the ground-truth heatmaps for joint \textit{j} of person 
\textit{i} at the current frame \textit{t}.


%% file: new_sec/4_result.tex
\section{Experiments}
\label{sec:experiment}

\subsection{Experimental Settings}
We evaluated our model on three widely used video benchmark datasets: PoseTrack2017~\cite{PoseTrack2017_CVPR2017}, PoseTrack2018~\cite{PoseTrack2018_CVPR2018}, and PoseTrack21~\cite{PoseTrack21_CVPR2022}. Each dataset contains dynamic video sequences with complex scenes, including severe occlusions, fast motions, and crowded environments. Performance was assessed using the Average Precision (AP) metric~\cite{HRNet_CVPR2019,PoseWarper_NIPS2019,DCPose_CVPR2021,RLE_ICCV2021}, where AP is computed for each keypoint and the mean Average Precision (mAP) is obtained by averaging AP across all keypoints. Each reported result is obtained  through 1–3 runs.
Our models were implemented in PyTorch. The encoder and decoder were pretrained on the COCO dataset~\cite{COCO_ECCV2014} following ViTPose~\cite{Vitpose_NIPS2022}, while the temporal modeling modules were randomly initialized and trained from scratch. The model was trained for 30 epochs on a single NVIDIA RTX A6000 GPU. The default mask threshold $\phi$ was set to 0.2.
Additional implementation details are provided in Appendix~A of the Supp. Material.

\subsection{Main Results}
\subsubsection{Comparison with the ViTPose Baseline}
To comprehensively assess the effectiveness of the proposed \mbox{TAR-ViTPose} on video inputs, we first compare it with the state-of-the-art ViT-based single-frame baseline, \textit{i.e.}, \mbox{ViTPose}~\cite{Vitpose_NIPS2022}. For a fair and thorough comparison, we evaluate both methods using four different scales of ViT backbones: ViT-S, ViT-B, ViT-L, and ViT-H, and ensure that each method employs the same pre-trained backbone and decoder. As reported in Table~\ref{tab:compare_to_vitpose_17}, our video-based approach consistently delivers substantial performance improvements across all backbones.
For instance, our method outperforms ViTPose~\cite{Vitpose_NIPS2022} by a notable margin of \textbf{2.3} mAP (or \textbf{2.1} mAP) when using the ViT-B (or ViT-H) backbone. In particular, by leveraging temporal information, our approach achieves significant improvements on challenging keypoints such as the wrist and ankle. For the ankle keypoint, for example, our method surpasses ViTPose~\cite{Vitpose_NIPS2022} by \textbf{3.8} mAP with the ViT-S backbone and by \textbf{3.4} mAP with the ViT-B backbone.
These results highlight the importance of leveraging temporal cues from adjacent frames, which the single-frame baseline (\textit{e.g.}, ViTPose) is inherently unable to capture.

\begin{figure}
    \centering
    \includegraphics[width=1.0\linewidth]{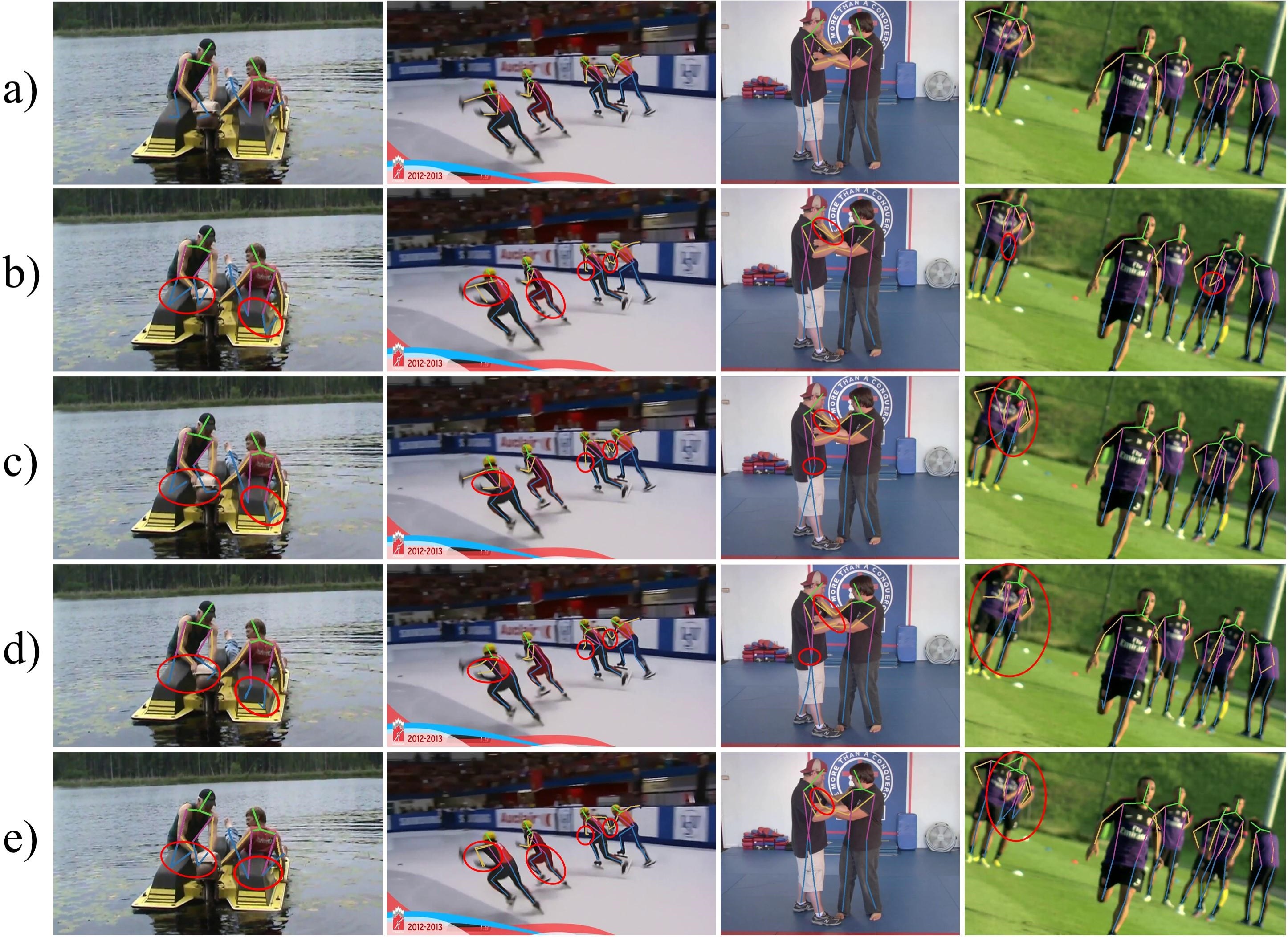}
    \caption{Qualitative comparison of a) our TAR-ViTPose, b) ViTPose \cite{Vitpose_NIPS2022}, c) DCPose \cite{DCPose_CVPR2021}, d) DSTA \cite{DSTA-CVPR2024}, and e) Poseidon \cite{Poseidon-Arxiv2025}, featuring challenges such as occlusion, motion blur, and defocus. The first two columns
are from the PoseTrack datasets, while the last two columns are
from in-the-wild videos. 
    Inaccurate predictions are marked with red solid circles. Zoom in for clarity.}
    \label{fig:qualitative_cmp}
    \vspace{-6mm}
\end{figure}

By leveraging temporal relationships between consecutive frames, our method demonstrates enhanced robustness in video scenes, particularly excelling in challenging scenarios such as occlusion and motion blur, as shown in Fig.~\ref{fig:qualitative_cmp}. These experiments validate that our video-based framework significantly outperforms the state-of-the-art ViT-based single-frame baseline ViTPose when processing video inputs, delivering superior overall performance.

\begin{table}[t]
\centering
\scriptsize
\setlength{\tabcolsep}{2.5pt}
\renewcommand\arraystretch{1.3}
\begin{tabular}{l|ccccccc|c}
\hline
Method    & Head          & Should.       & Elbow         & Wrist         & Hip           & Knee          & Ankle         & Mean \\ 
\hline
\multicolumn{9}{l}{\textit{Backbone}: ViT-S}    \\
ViTPose~\cite{Vitpose_NIPS2022}      & 86.5          & 87.2          & 81.5          & 75.0          & 78.9          & 78.4          & 70.4          & 80.1          \\
\rowcolor[gray]{0.9}
\textbf{TAR-ViTPose (Ours)} & \textbf{87.1}          & \textbf{88.0}          & \textbf{83.2}          & \textbf{77.8}          & \textbf{80.4}          & \textbf{81.0}          & \textbf{74.2}          & \textbf{81.9}          \\
\hline
\multicolumn{9}{l}{\textit{Backbone}: ViT-B}    \\
ViTPose~\cite{Vitpose_NIPS2022}    & 86.7          & 87.9          & 83.0          & 77.7          & 80.3          & 81.0          & 73.9          & 81.7          \\
\rowcolor[gray]{0.9}
\textbf{TAR-ViTPose (Ours)}   & \textbf{88.0}          & \textbf{89.2}          &\textbf{85.0}          & \textbf{80.3}         & \textbf{82.9}          & \textbf{84.2}          & \textbf{77.3}          & \textbf{84.0}          \\
\hline
\multicolumn{9}{l}{\textit{Backbone}: ViT-L}    \\
ViTPose~\cite{Vitpose_NIPS2022}  & 87.1          & 88.5          & 84.2          & 79.7          & 82.4          & 83.6          & 76.9          & 83.4          \\
\rowcolor[gray]{0.9}
\textbf{TAR-ViTPose (Ours)}   & \textbf{89.0}          & \textbf{90.4}         & \textbf{87.1}         & \textbf{82.3}          & \textbf{83.6}          & \textbf{85.8}          & \textbf{79.1}          & \textbf{85.3}          \\
\hline
\multicolumn{9}{l}{\textit{Backbone}: ViT-H}    \\
ViTPose~\cite{Vitpose_NIPS2022}    & 88.2          & 89.7          & 86.3          & 81.6          & 83.0          & 84.9          & 77.8          & 84.7          \\ 
\rowcolor[gray]{0.9}
\textbf{TAR-ViTPose (Ours)}   & \textbf{90.5} & \textbf{91.9} & \textbf{88.4} & \textbf{83.8} & \textbf{84.9} & \textbf{87.3} & \textbf{80.2} & \textbf{86.8} \\ 
\hline
\end{tabular}
\vspace{-2mm}
\caption{Comparison with the ViTPose baseline (mAP) on PoseTrack2017 val. set.}
\label{tab:compare_to_vitpose_17}
\vspace{-2mm}
\end{table}

\begin{table}[t]
\centering
\scriptsize
\setlength{\tabcolsep}{0.4mm}
\renewcommand\arraystretch{1.3}
\begin{tabular}{lccccccccc}
\hline
\multicolumn{1}{l|}{Method}                       & \multicolumn{1}{l|}{Backbone}   & Head          & Should.        & Elbow         & Wrist         & Hip           & Knee          & \multicolumn{1}{c|}{Ankle}         & Mean          \\ \hline
\multicolumn{10}{l}{\textit{Using bounding boxes predicted by the Faster R-CNN detector~\cite{FastRCNN-NIPS2015}}}                                                                                                                                                                              \\
\multicolumn{1}{l|}{PoseTrack \cite{PoseTrack_CVPR2018}}                    & \multicolumn{1}{c|}{ResNet-101} & 67.5          & 70.2          & 62.0          & 51.7          & 60.7          & 58.7          & \multicolumn{1}{c|}{49.8}          & 60.6          \\
\multicolumn{1}{l|}{PoseFlow \cite{PoseFlow-BMVC2018}}                     & \multicolumn{1}{c|}{ResNet-152} & 66.7          & 73.3          & 68.3          & 61.1          & 67.5          & 67.0          & \multicolumn{1}{c|}{61.3}          & 66.5          \\
\multicolumn{1}{l|}{SimBase. \cite{SimplePose_ECCV2018}}                     & \multicolumn{1}{c|}{ResNet-152} & 81.7          & 83.4          & 80.0          & 72.4          & 75.3          & 74.8          & \multicolumn{1}{c|}{67.1}          & 76.7          \\
\multicolumn{1}{l|}{HRNet \cite{HRNet_CVPR2019}}                        & \multicolumn{1}{c|}{HRNet-W48}  & 82.1          & 83.6          & 80.4          & 73.3          & 75.5          & 75.3          & \multicolumn{1}{c|}{68.5}          & 77.3          \\
\multicolumn{1}{l|}{PoseWarp. \cite{PoseWarper_NIPS2019}}                    & \multicolumn{1}{c|}{HRNet-W48}  & 81.4          & 88.3          & 83.9          & 78.0          & 82.4          & 80.5          & \multicolumn{1}{c|}{73.6}          & 81.2          \\
\multicolumn{1}{l|}{PAVE-Net$\dagger$ \cite{PAVENet_AAAI2026}}                    & \multicolumn{1}{c|}{Swin-L}  & 88.2          & 89.1          & 81.7          & 74.8          & 81.6          & 78.5          & \multicolumn{1}{c|}{71.8}          & 81.3          \\
\multicolumn{1}{l|}{DCPose \cite{DCPose_CVPR2021}}                       & \multicolumn{1}{c|}{HRNet-W48}  & 88.0          & 88.7          & 84.1          & 78.4          & 83.0          & 81.4          & \multicolumn{1}{c|}{74.2}          & 82.8          \\
\multicolumn{1}{l|}{FAMI-Pose \cite{FAMIPose_CVPR2022}}                    & \multicolumn{1}{c|}{HRNet-W48}  & 89.6          & 90.1          & 86.3          & 80.0          & 84.6          & 83.4          & \multicolumn{1}{c|}{77.0}          & 84.8          \\
\multicolumn{1}{l|}{DSTA \cite{DSTA-CVPR2024}}                         & \multicolumn{1}{c|}{HRNet-W48}      & 89.8          & 90.8          & 86.2          & 79.3          & 85.2          & 82.2          & \multicolumn{1}{c|}{75.9}          & 84.6          \\
\multicolumn{1}{l|}{DSTA \cite{DSTA-CVPR2024}}                         & \multicolumn{1}{c|}{ViT-H}      & 89.3          & 90.6          & 87.3          & 82.6          & 84.5          & 85.1          & \multicolumn{1}{c|}{77.8}          & 85.6          \\
\rowcolor[gray]{0.9}
\multicolumn{1}{l|}{\textbf{TAR-ViTPose}}         & \multicolumn{1}{c|}{ViT-S}      & 87.1          & 88.0          & 83.2          & 77.8          & 80.4          & 81.0          & \multicolumn{1}{c|}{74.2}          & 81.9          \\
\rowcolor[gray]{0.9}
\multicolumn{1}{l|}{\textbf{TAR-ViTPose}}         & \multicolumn{1}{c|}{ViT-B}      & 88.0          & 89.2          & 85.0          & 80.3          & 82.9          & 84.2          & \multicolumn{1}{c|}{77.3}          & 84.0          \\
\rowcolor[gray]{0.9}
\multicolumn{1}{l|}{\textbf{TAR-ViTPose}}   & \multicolumn{1}{c|}{ViT-H}      & \textbf{90.5} & \textbf{91.9} & \textbf{88.4} & \textbf{83.8} & \textbf{84.9} & \textbf{87.3} & \multicolumn{1}{c|}{\textbf{80.2}} & \textbf{86.8} \\ \hline
\multicolumn{10}{l}{\textit{Using bounding boxes from unspecified sources or ground-truth ones (*)}}\\
\multicolumn{1}{l|}{STEmbed \cite{STEmbeding_CVPR2019}}                     & \multicolumn{1}{c|}{ResNet-152} & 83.8          & 81.6          & 77.1          & 70.0          & 77.4          & 74.5          & \multicolumn{1}{c|}{70.8}          & 77.0          \\
\multicolumn{1}{l|}{Dyn.-GNN \cite{DynGNN_CVPR2021}}                    & \multicolumn{1}{c|}{HRNet-W48}  & 88.4          & 88.4          & 82.0          & 74.5          & 79.1          & 78.3          & \multicolumn{1}{c|}{73.1}          & 81.1          \\
\multicolumn{1}{l|}{DetTrack \cite{DetTrack_CVPR2020}}                    & \multicolumn{1}{c|}{HRNet-W48}  & 89.4          & 89.7          & 85.5          & 79.5          & 82.4          & 80.8          & \multicolumn{1}{c|}{76.4}          & 83.8          \\
\multicolumn{1}{l|}{TDMI \cite{TDMI_CVPR2023}}                        & \multicolumn{1}{c|}{HRNet-W48}  & 90.6          & 91.0          & 87.2          & 81.5          & 85.2          & 84.5          & \multicolumn{1}{c|}{78.7}          & 85.9          \\
\multicolumn{1}{l|}{DiffPose \cite{DiffPose-ICCV2023}}                     & \multicolumn{1}{c|}{ViT (\#)}       & 89.0          & 91.2          & 87.4          & 83.5          & 85.5          & 87.2          & \multicolumn{1}{c|}{80.2}          & 86.4          \\
\multicolumn{1}{l|}{CM-Pose \cite{CMPose-AAAI2025}}                     & \multicolumn{1}{c|}{ViT (\#)}        & 89.2          & 92.0          & 89.0          & 85.6          & \textbf{88.6} & 87.2          & \multicolumn{1}{c|}{81.1}          & 87.5          \\ 
\multicolumn{1}{l|}{GLSMamba \cite{HRSM-ICCV2025}}                  & \multicolumn{1}{c|}{ViT-H}        & 90.7        & 92.1          & 89.2          & 85.3          & 87.0         & 88.4          & \multicolumn{1}{c|}{82.4}          & 88.0          \\ 
\multicolumn{1}{l|}{Poseidon*  \cite{Poseidon-Arxiv2025}}                  & \multicolumn{1}{c|}{ViT-H}        & 91.9        & 93.1          & 88.6          & 85.7          & 88.2         & 89.6          & \multicolumn{1}{c|}{84.4}          & 88.9          \\ 
\rowcolor[gray]{0.9}
\multicolumn{1}{l|}{\textbf{TAR-ViTPose*}}      & \multicolumn{1}{c|}{ViT-S}        & 88.0        & 88.9          & 84.2          & 78.4          & 82.6         & 81.5          & \multicolumn{1}{c|}{76.9}          & 83.2          \\
\rowcolor[gray]{0.9}
\multicolumn{1}{l|}{\textbf{TAR-ViTPose*}}      & \multicolumn{1}{c|}{ViT-B}        & 92.3        & 92.6          & 88.2          & 83.3          & 85.9         & 87.4          & \multicolumn{1}{c|}{82.2}          & 87.7          \\
\rowcolor[gray]{0.9}
\multicolumn{1}{l|}{\textbf{TAR-ViTPose*}}      & \multicolumn{1}{c|}{ViT-H}      & \textbf{93.3} & \textbf{94.4} & \textbf{90.6} & \textbf{87.0} & 86.5          & \textbf{90.0} & \multicolumn{1}{c|}{\textbf{88.9}} & \textbf{90.3} \\ \hline
\end{tabular}
\vspace{-2mm}
\caption{Comparison with the SOTAs on PoseTrack2017 val. set.  `$\dagger$' indicates an end-to-end method without bounding box detection, and `\#' indicates that the ViT backbone used is not specified. Similar to FAMI-Pose~\cite{FAMIPose_CVPR2022} and DSTA~\cite{DSTA-CVPR2024}, our proposed TAR-ViTPose sets the temporal span $T$ to 2, which includes two preceding and two succeeding frames, totaling four auxiliary frames.}
\label{tab:compare_to_sota_17}
\vspace{-6mm}
\end{table}

\subsubsection{Comparison with State-of-the-Art Methods}
We begin with a comprehensive performance comparison against state-of-the-art methods on the PoseTrack2017 dataset and further evaluate on PoseTrack2018 and PoseTrack21. Similar to our approach, most existing video-based human pose estimation methods adopt a two-stage top-down framework, where the human detection results in the first stage have a substantial impact on the pose estimation performance in the second stage. However, we observe that some state-of-the-art methods report improved pose estimation accuracy by using ground-truth bounding boxes instead of those predicted by human detectors such as Faster R-CNN~\cite{FastRCNN-NIPS2015}. In addition, several methods do not clearly specify the source of their bounding boxes in their papers, and their source code is not publicly available, making it difficult to determine whether they rely on ground-truth or predicted bounding boxes. To ensure a fair comparison, we explicitly mark these cases in the tables.

\textbf{Results on the PoseTrack Datasets.} 
Table~\ref{tab:compare_to_sota_17} presents the quantitative results of different approaches on the PoseTrack2017 validation set.
Compared with existing state-of-the-art methods that also use bounding boxes predicted by the human detector~\cite{FastRCNN-NIPS2015}, our method achieves performance on par with the best-performing approaches. For example, by fully leveraging the strong global modeling capacity of ViTs, our method attains an impressive 81.9 mAP even with the small-scale ViT-S backbone, outperforming the much larger backbone network HRNet-W48~\cite{HRNet_CVPR2019} by \textbf{4.6} points.
Notably, when equipped with the same ViT-H backbone, our method further surpasses the current state-of-the-art DSTA by \textbf{1.2} mAP, setting a new state-of-the-art performance of \textbf{86.8} mAP. Performance gains on relatively challenging joints are particularly encouraging: 
the ankle reaches 80.2 mAP ($\uparrow$ \textbf{2.4}), 
the wrist reaches 83.8 mAP ($\uparrow$ \textbf{1.2}), 
and the knee reaches 87.3 mAP ($\uparrow$ \textbf{2.2}). 
It is also worth noting that methods leveraging temporal information, such as 
PoseWarper~\cite{PoseWarper_NIPS2019}, DCPose~\cite{DCPose_CVPR2021}, FAMI-Pose~\cite{FAMIPose_CVPR2022}, 
DSTA~\cite{DSTA-CVPR2024}, and our TAR-ViTPose, consistently outperform those relying solely on a single key frame, such as SimpleBaseline~\cite{SimplePose_ECCV2018} and HRNet~\cite{HRNet_CVPR2019}. 
This further reinforces the importance of incorporating temporal cues from adjacent frames. 
Qualitative results are provided in Fig.~\ref{fig:qualitative_cmp}.

To explore the upper-bound performance of our proposed TAR-ViTPose, we use ground-truth bounding boxes instead of detector-predicted ones to localize individuals in the video frames. As shown in Table~\ref{tab:compare_to_sota_17}, our method achieves substantial performance gains across ViT backbones of different scales. Taking ViT-H as an example, our approach yields an additional improvement of 3.5 mAP, reaching an impressive \textbf{90.3} mAP, surpassing the state-of-the-art video-based method Poseidon~\cite{Poseidon-Arxiv2025}—which also uses ground-truth bounding boxes—by \textbf{1.4} mAP. These results clearly demonstrate the effectiveness of our proposed joint-centric temporal aggregation and restoring mechanisms.

We further evaluate our model on the PoseTrack2018 and PoseTrack21 datasets. Due to space limitations, these results are reported in Appendix B of the supplementary materials. Based on the evaluations, our approach achieves new state-of-the-art performance, obtaining \textbf{84.2} mAP and \textbf{84.1} mAP on the two datasets, respectively. When evaluated using ground-truth bounding boxes, our method further reaches 89.8 mAP and 91.0 mAP, yielding additional improvements of \textbf{5.6} and \textbf{6.9} points, respectively.

\begin{table}[t]
\centering
\small
\begin{tabular}{l|c|ccc}
\hline
Method                       & Backbone  & \#Params & FPS(↑) &  mAP        \\ \hline
PoseWarper \cite{PoseWarper_NIPS2019}                   & HRNet-W48 & 71.1M    & 52 & 81.0             \\
DCPose \cite{DCPose_CVPR2021}                       & HRNet-W48 & 65.2M    & 128 & 82.8           \\
DSTA \cite{DSTA-CVPR2024}                         & HRNet-W48 & 63.7M    & 156 &  83.4           \\
DSTA \cite{DSTA-CVPR2024}                         & ViT-H     & 422.2M   & 25  &  84.3           \\ 
\rowcolor[gray]{0.9}
 & ViT-S     & 35.6M    & \textbf{413} & 81.5   \\
\rowcolor[gray]{0.9}
                             & ViT-B     & 101.1M   & {186} &  83.4  \\
\rowcolor[gray]{0.9}                             
\multirow{-3}{*}{TAR-ViTPose}                             & ViT-H     & 672.5M   & {28}  & \textbf{86.3}           \\ \hline
\end{tabular}
\caption{Runtime frame rate (FPS), measured on an A6000. All methods
utilize the same two auxiliary frames as in~\cite{DCPose_CVPR2021}.}
\label{tab:compare_to_sota_fps}
\vspace{-6mm}
\end{table}

\textbf{Comparison of Runtime Frame Rates.} 
We conduct experiments to compare the runtime frame rates of various methods in real-world video applications. For a fair comparison, we use the official open-source implementations of existing state-of-the-art video-based approaches including PoseWarper~\cite{PoseWarper_NIPS2019}, DCPose~\cite{DCPose_CVPR2021}, and DSTA~\cite{DSTA-CVPR2024}, and employ two auxiliary frames for all methods. 
Table~\ref{tab:compare_to_sota_fps} summarizes the resulting frame rates (FPS). The speed of all methods is measured on a single A6000 GPU with a batch size of 16. 
During inference, we adopt a simple bounding box intersection over union (IoU) strategy for human tracking across frames. This design ensures that each cropped human image undergoes backbone processing only once, and the extracted features are then reused for pose estimation in the current, preceding, and succeeding frames.

As observed, when using the small ViT-S backbone, our TAR-ViTPose achieves an impressive \textbf{413} FPS, significantly outperforming PoseWarper~\cite{PoseWarper_NIPS2019} and DCPose~\cite{DCPose_CVPR2021} while delivering higher or comparable accuracy. Furthermore, DSTA~\cite{DSTA-CVPR2024}, being a regression-based method, requires fewer parameters and less computation than heatmap-based approaches, which contributes to its high frame rate. Thanks to our design strategy, TAR-ViTPose, like ViTPose~\cite{Vitpose_NIPS2022}, fully preserves the plain ViT architecture and adopts a lightweight decoding pipeline. Although the model size of TAR-ViTPose becomes relatively large when using ViT-B and ViT-H backbones, it achieves an excellent balance between throughput and accuracy, delivering not only the highest accuracy but also a higher frame rate than DSTA. These results demonstrate that the plain vision transformer possesses strong representational capability and is well aligned with modern hardware.

\subsection{Ablation Study}
We conduct ablation experiments on the PoseTrack2017 validation set. The temporal span \(T\) is set to 2, consisting of two preceding and two subsequent frames, totalling 4 auxiliary frames, and the ViT-B backbone is employed.

\begin{table}[t]
\centering
\begin{tabular}{l|cc}
\hline
Method                                 & GFLOPs             & mAP           \\ \hline
(a) Self-attention (All frames)        & 38.22              & 82.2          \\
(b) Cross-attention (Cur and Aux)      & 12.74              & 82.6          \\
(c) Self-attention (Aux) + (b)         & 16.99              & 82.8          \\ 
(d) Our joint-specific modeling        & 3.89               & \textbf{84.0} \\
\hline
\end{tabular}
\vspace{-2mm}
\caption{Ablation of different temporal modeling strategies.}
\label{tab:diff_temporal_modeling}
\vspace{-2mm}
\end{table}

\textbf{Choice of Temporal Modeling Strategy.} We examine the effectiveness of our joint-specific temporal modeling strategy by comparing it with various alternative designs. Several straightforward self-attention–based methods~\cite{SelfAttention-NIPS2017} can be used to aggregate auxiliary-frame features to enhance the current-frame representation:
(\textbf{a}) directly performing self-attention over the feature tokens of all frames;
(\textbf{b}) applying cross-attention between the current frame's feature tokens (as queries) and those of the auxiliary frames (as keys and values);
(\textbf{c}) first performing self-attention over the auxiliary-frame tokens, followed by the cross-attention strategy described in (b), as adopted in~\cite{Poseidon-Arxiv2025}.
As shown in Table~\ref{tab:diff_temporal_modeling}, our joint-specific temporal modeling strategy effectively improves accuracy by \textbf{1.2} points while incurring lower computational cost. This confirms the importance of accurately aligning temporally corresponding keypoint features across frames within the temporal modeling module.

\begin{table}[t]
\centering
\footnotesize
\begin{tabular}{c|clcl|c}
\multirow{3}{*}{Method} & \multicolumn{2}{c|}{\textbf{JTA}}                                   & \multicolumn{2}{c|}{\textbf{GRA}}                                 & \multirow{3}{*}{mAP} \\
                        & \multicolumn{1}{l}{\#Params} & \multicolumn{1}{l|}{GFLOPs} & \multicolumn{1}{l}{\#Params} & MFLOPs                    &                      \\
                        & 14.19M                       & \multicolumn{1}{c|}{3.87}   & 2.4M                         & \multicolumn{1}{c|}{21.23} &                      \\ \hline
(a)                     & \multicolumn{2}{c}{\checkmark}                                      & \multicolumn{2}{c|}{\ding{55}}                                   & 70.3                 \\
(b)                     & \multicolumn{2}{c}{\checkmark}                                      & \multicolumn{2}{c|}{\checkmark}                                  & \textbf{84.0}          \\ 
\hline
\multicolumn{5}{l|}{ViTPose baseline~\cite{Vitpose_NIPS2022}} & 81.7 \\
\hline
\end{tabular}
\vspace{-2mm}
\caption{Effect of JTA and GRA in our temporal modeling.}
\label{tab:diff_module}
\vspace{-6mm}
\end{table}

\textbf{Effect of JTA and GRA.} 
As shown in Table~\ref{tab:diff_module}, within our joint-specific temporal modeling framework, it is essential to reinject the aggregated temporal features produced by JTA back into the current-frame features through GRA. This step enriches the representation while preserving the global context required for precise keypoint localization. In contrast, directly regressing keypoints from the joint query tokens that contain temporal information, following~\cite{DSTA-CVPR2024}, results in a substantial performance drop of up to \textbf{13.7} points—even falling noticeably below the single-frame ViTPose baseline—because these query tokens lack the global context among keypoints.
Notably, both JTA and GRA introduce only a minimal number of additional parameters and incur very low computational overhead, making them lightweight and suitable for real-time applications.

\begin{figure}
    \centering
    \includegraphics[width=1.0\linewidth]{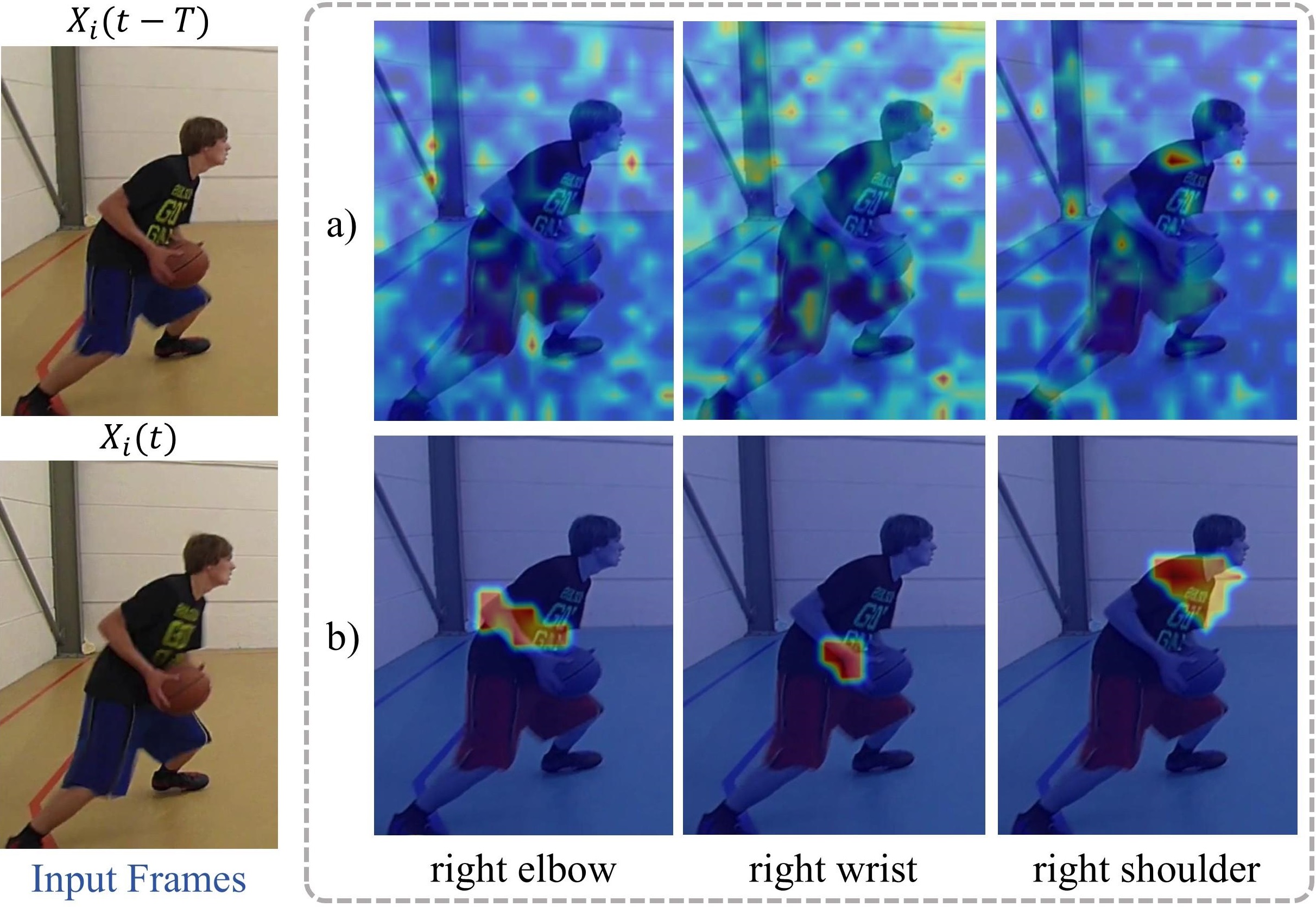}
    \vspace{-6mm}
    \caption{Visualization of attention heatmaps for joint query tokens with (b) and without (a) mask-aware attention. Given a current frame $X_i(t)$ and a neighboring frame $X_i(t-T)$, we visualize the attention heatmaps of three different joint query tokens with respect to the features of $X_i(t-T)$. See Supp. Material for more.}
    \label{fig:query_vis}
    \vspace{-2mm}
\end{figure}

\begin{table}[t]
\centering
\footnotesize
\renewcommand\arraystretch{1.2}
\begin{tabular}{c|c|c}
\hline
Method & Masking         & mAP              \\ \hline
(a)    & \ding{55}       & 82.6             \\
(b)    & \checkmark      & \textbf{84.0}    \\ \hline
\end{tabular}
\vspace{-2mm}
\caption{Impact of mask-aware attention.}
\label{tab:masking_tab}
\vspace{-6mm}
\end{table}

\textbf{Impact of Mask-aware Attention.} 
To verify that the proposed mask-aware attention enables each joint query token to effectively focus on its corresponding keypoint regions across frames while avoiding interference from unrelated areas, we visualize the attention heatmaps of several joint query tokens, including the right elbow, right wrist, and right shoulder, as shown in Fig.~\ref{fig:query_vis}. Without mask-aware attention, the attention regions of each joint query token are highly scattered and may even extend into background areas. In contrast, with mask-aware attention, the attention of each joint query token is concentrated around its corresponding keypoint region, enabling accurate aggregation of temporally coherent features for each keypoint. As a result, mask-aware attention improves the accuracy by an additional \textbf{1.4} mAP, as shown in Table~\ref{tab:masking_tab}.

%% file: new_sec/X_suppl.tex
\clearpage
\setcounter{page}{1}
\maketitlesupplementary

\section*{Appendix}
\label{sec:appendix}

In the supplementary material, we provide:
\begin{description}
    \item[\S\textcolor{red}{A}] Additional Implementation Details.
    \item[\S\textcolor{red}{B}] Experiments on PoseTrack2018/21 Datasets.
    \item[\S\textcolor{red}{C}] Additional Ablation Study.
    \item[\S\textcolor{red}{D}] Qualitative Results.
    \item[\S\textcolor{red}{E}] Limitation and Future Work.
\end{description}

\section*{A. Additional Implementation Details}
\label{sec:appendix_impl_details}

\noindent\textbf{Dataset.}
Our models are evaluated on three widely used video-based human pose estimation benchmarks: PoseTrack2017~\cite{PoseTrack2017_CVPR2017}, PoseTrack2018~\cite{PoseTrack2018_CVPR2018}, and PoseTrack21~\cite{PoseTrack21_CVPR2022}. Together, these datasets form a comprehensive evaluation suite that spans diverse scenarios, motion patterns, and levels of visual complexity. A brief overview of each dataset is provided below.

\begin{itemize}
\item \textbf{PoseTrack2017}~\cite{PoseTrack2017_CVPR2017} contains 250 training videos and 50 validation videos, with a total of 80,144 pose annotations. Each person instance is annotated with 15 keypoints and corresponding visibility labels. Training clips are densely annotated in the central 30 frames, whereas validation clips are annotated every four frames.
\item \textbf{PoseTrack2018}~\cite{PoseTrack2018_CVPR2018} substantially enlarges the dataset, providing 593 training videos and 170 validation videos, amounting to 153,615 pose annotations. It uses the same 15-keypoint annotation scheme and visibility labels as PoseTrack2017, with an identical annotation protocol—dense annotations in the central 30 frames of training videos and sparser annotations (every four frames) for validation videos.

\item \textbf{PoseTrack21}~\cite{PoseTrack21_CVPR2022} extends and refines PoseTrack2018, particularly improving annotations for small individuals and people in crowded scenes. It includes 177,164 human pose annotations while preserving the same keypoint definition and annotation strategy. The updates make it more challenging, especially regarding occlusion and scale variations.
\end{itemize}

\noindent\textbf{Optimization.} 
We implemented TAR-ViTPose in PyTorch. Data augmentation includes random scaling within the range \([0.65, 1.35]\), random rotation within \([-45^\circ, 45^\circ]\), random cropping, and horizontal flipping. The input image resolution is set to \(384\times288\). The initial learning rate is \(5\times10^{-6}\) and is reduced by 50\% every 5 epochs. We use the AdamW optimizer for model training.

\section*{B. Experiments on PoseTrack2018/21 Datasets}
\subsection*{B1. Comparison with the ViTPose Baseline}

\begin{table}[t]
\centering
\scriptsize
\setlength{\tabcolsep}{2.5pt}
\renewcommand\arraystretch{1.3}
\begin{tabular}{l|ccccccc|c}
\hline
Method                     & Head          & Should.        & Elbow         & Wrist         & Hip           & Knee          & Ankle         & Mean          \\
\hline
\multicolumn{9}{l}{\textit{Backbone}: ViT-S}                                                                                                                        \\
ViTPose \cite{Vitpose_NIPS2022}                    & 80.6          & 84.1          & 79.9          & 73.5          & 77.0          & 76.2          & 70.3          & 78.2          \\
\rowcolor[gray]{0.9} 
\textbf{TAR-ViTPose(Ours)} & \textbf{82.8} & \textbf{85.6} & \textbf{81.5} & \textbf{75.2} & \textbf{78.0} & \textbf{78.1} & \textbf{72.1} & \textbf{79.3} \\
\hline
\multicolumn{9}{l}{\textit{Backbone}: ViT-B}                                                                                                                        \\
ViTPose \cite{Vitpose_NIPS2022}                    & 83.2          & 86.3          & 82.8          & 77.5          & 78.2          & 79.4          & 74.9          & 80.5          \\
\rowcolor[gray]{0.9} 
\textbf{TAR-ViTPose(Ours)} & \textbf{84.3} & \textbf{87.8} & \textbf{83.6} & \textbf{79.4} & \textbf{79.3} & \textbf{82.8} & \textbf{76.5} & \textbf{82.1} \\
\hline
\multicolumn{9}{l}{\textit{Backbone}: ViT-H}                                                                                                                        \\
ViTPose \cite{Vitpose_NIPS2022}                    & 85.4          & 87.0          & 84.7          & 81.0          & 79.1          & 82.0          & 76.5          & 82.4          \\
\rowcolor[gray]{0.9} 
\textbf{TAR-ViTPose(Ours)} & \textbf{86.8} & \textbf{88.9} & \textbf{86.1} & \textbf{81.9} & \textbf{82.9} & \textbf{83.4} & \textbf{77.9} & \textbf{84.2} \\
\hline
\end{tabular}
\caption{Comparison with the ViTPose baseline (mAP) on Pose-
Track2018 val. set.}
\label{tab:compare_to_vitpose_18}
\end{table}

\begin{table}[t]
\centering
\scriptsize
\setlength{\tabcolsep}{2.5pt}
\renewcommand\arraystretch{1.3}
\begin{tabular}{l|ccccccc|c}
\hline
Method                     & Head          & Should.        & Elbow         & Wrist         & Hip           & Knee          & Ankle         & Mean          \\
\hline
\multicolumn{9}{l}{\textit{Backbone}: ViT-S}                                                                                                                        \\
ViTPose \cite{Vitpose_NIPS2022}                    & 79.7          & 84.4          & 80.4          & 73.8          & 77.2          & 77.3          & 71.2          & 77.8          \\
\rowcolor[gray]{0.9} 
\textbf{TAR-ViTPose(Ours)} & \textbf{81.6} & \textbf{84.7} & \textbf{80.9} & \textbf{74.2} & \textbf{77.8} & \textbf{77.9} & \textbf{71.7} & \textbf{78.5} \\
\hline
\multicolumn{9}{l}{\textit{Backbone}: ViT-B}                                                                                                                        \\
ViTPose \cite{Vitpose_NIPS2022}                    & 81.8          & 85.2          & 82.1          & 76.6          & 78.5          & 79.5          & 74.5          & 79.9          \\
\rowcolor[gray]{0.9} 
\textbf{TAR-ViTPose(Ours)} & \textbf{82.7} & \textbf{85.9} & \textbf{83.5} & \textbf{79.2} & \textbf{79.5} & \textbf{81.7} & \textbf{76.8} & \textbf{81.4} \\
\hline
\multicolumn{9}{l}{\textit{Backbone}: ViT-H}                                                                                                                        \\
ViTPose \cite{Vitpose_NIPS2022}                    & 83.2          & 87.0          & 84.1          & 80.4          & 79.5          & 81.9          & 77.3          & 82.0          \\
\rowcolor[gray]{0.9} 
\textbf{TAR-ViTPose(Ours)} & \textbf{86.6} & \textbf{88.3} & \textbf{85.3} & \textbf{82.0} & \textbf{83.7} & \textbf{83.4} & \textbf{78.2} & \textbf{84.1} \\
\hline
\end{tabular}
\caption{Comparison with the ViTPose baseline (mAP) on Pose-
Track21 val. set.}
\label{tab:compare_to_vitpose_21}
\vspace{-4mm}
\end{table}

Tables~\ref{tab:compare_to_vitpose_18} and \ref{tab:compare_to_vitpose_21} present comparisons between our method and the state-of-the-art ViT-based single-frame baseline ViTPose~\cite{Vitpose_NIPS2022} on the PoseTrack2018 and PoseTrack21 validation sets, respectively. The results further demonstrate that our video-based approach consistently achieves substantial performance gains across all ViT backbones, highlighting the importance of exploiting temporal cues from adjacent frames, which single-frame baselines such as ViTPose are inherently unable to capture.

\subsection*{B2. Comparison with State-of-the-Art Methods}
As shown in Tables~\ref{tab:compare_to_sota_18} and \ref{tab:compare_to_sota_21}, our approach establishes new state-of-the-art performance on the PoseTrack2018 and PoseTrack21 datasets, achieving \textbf{84.2} mAP and \textbf{84.1} mAP, respectively. When evaluated with ground-truth bounding boxes, our method attains 89.8 mAP and 91.0 mAP, yielding additional gains of \textbf{5.6} and \textbf{6.9} points. These results consistently outperform the state-of-the-art video-based methods MTPose~\cite{MTPose-ijcai2024} and Poseidon~\cite{Poseidon-Arxiv2025}, both of which also rely on ground-truth bounding boxes.

\begin{table}[t]
\centering
\scriptsize
\setlength{\tabcolsep}{0.4mm}
\renewcommand\arraystretch{1.3}
\begin{tabular}{l|c|ccccccc|c}
\hline
Method                & \multicolumn{1}{l|}{Backbone}  & Head          & Should.        & Elbow         & Wrist         & Hip           & Knee          & Ankle         & Mean          \\
\hline
\multicolumn{10}{l}{\textit{Using bounding boxes predicted by the Faster R-CNN detector} \cite{FastRCNN-NIPS2015}}                                                                          \\
PoseWarper \cite{PoseWarper_NIPS2019}            & HRNet-W48 & 79.9          & 86.3          & 82.4          & 77.5          & 79.8          & 78.8          & 73.2          & 79.7          \\
PAVE-Net$\dagger$ \cite{PAVENet_AAAI2026}            & Swin-L & 84.9          & 87.5          & 81.7          & 74.5          & 77.9          & 76.9          & 72.4          & 79.7          \\
DCPose \cite{DCPose_CVPR2021}                & HRNet-W48 & 84.0          & 86.6          & 82.7          & 78.0          & 80.4          & 79.3          & 73.8          & 80.9          \\
FAMI-Pose \cite{FAMIPose_CVPR2022}             & HRNet-W48 & 85.5          & 87.7          & 84.2          & 79.2          & 81.4          & 81.1          & 74.9          & 82.2          \\
DSTA \cite{DSTA-CVPR2024}                  & ViT-H     & 85.9          & 88.8          & 85.0          & 81.1          & 81.5          & 83.0          & 77.4          & 83.4          \\
\rowcolor[gray]{0.9}
\textbf{TAR-ViTPose}  & ViT-S     & 82.8          & 85.6          & 81.5          & 75.2          & 78.0          & 78.1          & 72.1          & 79.3          \\
\rowcolor[gray]{0.9}
\textbf{TAR-ViTPose}  & ViT-B     & 84.3          & 87.8          & 83.6          & 79.4          & 79.3          & 82.8          & 76.5          & 82.1          \\
\rowcolor[gray]{0.9}
\textbf{TAR-ViTPose}  & ViT-H     & \textbf{86.8} & \textbf{88.9} & \textbf{86.1} & \textbf{81.9} & \textbf{82.9} & \textbf{83.4} & \textbf{77.9} & \textbf{84.2} \\
\hline
\multicolumn{10}{l}{\textit{Using bounding boxes from unspecified sources or ground-truth ones (*)}}                                                              \\
Dyn.-GNN \cite{DynGNN_CVPR2021}              & HRNet-W48 & 80.6          & 84.5          & 80.6          & 74.4          & 75.0          & 76.7          & 71.8          & 77.9          \\
DetTrack \cite{DetTrack_CVPR2020}              & HRNet-W48 & 84.9          & 87.4          & 84.8          & 79.2          & 77.6          & 79.7          & 75.3          & 81.5          \\
TDMI \cite{TDMI_CVPR2023}                  & HRNet-W48 & 86.2          & 88.7          & 85.4          & 80.6          & 82.4          & 82.1          & 77.5          & 83.5          \\
DiffPose \cite{DiffPose-ICCV2023}              & ViT (\#)  & 85.0          & 87.7          & 84.3          & 81.5          & 81.4          & 82.9          & 77.6          & 83.0          \\
MTPose* \cite{MTPose-ijcai2024}               & ViT (\#)       & 89.4          & 92.4          & 90.1          & 87.3          & \textbf{85.7} & 89.7          & 88.1          & 89.0          \\
CM-Pose \cite{CMPose-AAAI2025}               & ViT (\#)  & 85.7          & 88.9          & 85.8          & 81.0          & 84.4          & 84.2          & 80.1          & 84.4          \\
GLSMamba \cite{HRSM-ICCV2025}            & ViT-H     & 85.6          & 88.9          & 86.5          & 83.6          & 82.9          & 85.7          & 81.4          & 84.9          \\
Poseidon* \cite{Poseidon-Arxiv2025}             & ViT-H     & 88.8          & 91.4          & 88.6          & 86.3          & 83.3          & 88.8          & 87.2          & 87.8          \\
\rowcolor[gray]{0.9}
\textbf{TAR-ViTPose*} & ViT-S     & 85.6          & 88.5          & 84.2          & 79.2          & 80.8          & 81.9          & 78.7          & 82.9          \\
\rowcolor[gray]{0.9}
\textbf{TAR-ViTPose*} & ViT-B     & 88.7          & 91.8          & 89.4          & 86.1          & 83.6          & 88.5          & 87.3          & 88.0          \\
\rowcolor[gray]{0.9}
\textbf{TAR-ViTPose*} & ViT-H     & \textbf{90.9} & \textbf{93.1} & \textbf{91.1} & \textbf{88.5} & 84.7          & \textbf{90.4} & \textbf{89.0} & \textbf{89.8} \\
\hline
\end{tabular}
\caption{Comparison with the SOTAs on PoseTrack2018 val. set. ‘\#’ indicates that the ViT backbone used is not specified, `$\dagger$' indicates the end-to-end approach. Similar to FAMI-Pose \cite{FAMIPose_CVPR2022} and DSTA \cite{DSTA-CVPR2024}, our proposed TAR-ViTPose sets the temporal span \(T\) to 2, which includes two preceding and two succeeding frames, totaling four auxiliary frames.}
\label{tab:compare_to_sota_18}
\end{table}

\begin{table}[t]
\centering
\scriptsize
\setlength{\tabcolsep}{0.4mm}
\renewcommand\arraystretch{1.3}
\begin{tabular}{l|c|ccccccc|c}
\hline
Method                & \multicolumn{1}{l|}{Backbone}  & Head          & Should.        & Elbow         & Wrist         & Hip           & Knee          & Ankle         & Mean          \\
\hline
\multicolumn{10}{l}{\textit{Using bounding boxes predicted by the Faster R-CNN detector} \cite{FastRCNN-NIPS2015}}                                                                          \\
SimBase. \cite{SimplePose_ECCV2018}              & ResNet-152                   & 80.5          & 81.2          & 73.2          & 64.8          & 73.9          & 72.7          & 67.7          & 73.9          \\
HRNet \cite{HRNet_CVPR2019}                 & HRNet-W48                    & 81.5          & 83.2          & 81.1          & 75.4          & 79.2          & 77.8          & 71.9          & 78.8          \\
PAVE-Net $\dagger$ \cite{PAVENet_AAAI2026}                 & Swin-L                    & 84.7          & 86.5          & 81.9          & 74.7          & 77.4          & 76.6          & 71.4          & 79.4          \\
PoseWarper \cite{PoseWarper_NIPS2019}            & HRNet-W48                    & 82.3          & 84.0          & 82.2          & 75.5          & 80.7          & 78.7          & 71.6          & 79.5          \\
DCPose \cite{DCPose_CVPR2021}                & HRNet-W48                    & 83.7          & 84.4          & 82.6          & 78.7          & 80.1          & 79.8          & 74.4          & 80.7          \\
FAMI-Pose \cite{FAMIPose_CVPR2022}             & HRNet-W48                    & 83.3          & 85.4          & 82.9          & 78.6          & 81.3          & 80.5          & 75.3          & 81.2          \\
DSTA \cite{DSTA-CVPR2024}                  & ViT-H                        & \textbf{87.5} & 87.0          & 84.2          & 81.4          & 82.3          & 82.5          & 77.7          & 83.5          \\
\rowcolor[gray]{0.9}
\textbf{TAR-ViTPose}  & ViT-S    & 79.0          & 83.1          & 79.8          & 73.6          & 75.3          & 77.2          & 72.6          & 77.4          \\
\rowcolor[gray]{0.9}
\textbf{TAR-ViTPose}  & ViT-B     & 82.7          & 85.9          & 83.5          & 79.2          & 79.5          & 81.7          & 76.8          & 81.4          \\
\rowcolor[gray]{0.9}
\textbf{TAR-ViTPose}  & ViT-H     & 86.6          & \textbf{88.3} & \textbf{85.3} & \textbf{82.0} & \textbf{83.7} & \textbf{83.4} & \textbf{78.2} & \textbf{84.1} \\
\hline
\multicolumn{10}{l}{\textit{Using bounding boxes from unspecified sources or ground-truth ones (*)}}                                                              \\
TDMI \cite{TDMI_CVPR2023}                  & HRNet-W48                    & 85.8          & 87.5          & 85.1          & 81.2          & 83.5          & 82.4          & 77.9          & 83.5          \\
DiffPose \cite{DiffPose-ICCV2023}              & ViT (\#)                     & 84.7          & 85.6          & 83.6          & 80.8          & 81.4          & 83.5          & 80.0          & 82.9          \\
MTPose* \cite{MTPose-ijcai2024}               & ViT (\#)                          & 92.0          & 91.7          & 88.7          & 85.5          & 86.4          & 86.6          & 85.3          & 88.3          \\
CM-Pose \cite{CMPose-AAAI2025}               & ViT (\#)                     & 88.9          & 88.3          & 84.4          & 81.9          & 84.6          & 83.7          & 78.8          & 84.3          \\
GLSMamba \cite{HRSM-ICCV2025}            & ViT-H                        & 87.0          & 86.9          & 85.4          & 83.2          & 83.4          & 84.8          & 80.8          & 84.7          \\
Poseidon* \cite{Poseidon-Arxiv2025}            & ViT-H                        & 92.2          & 90.8          & 88.3          & 85.8          & 85.5          & 87.7          & 85.7          & 88.3          \\
\rowcolor[gray]{0.9}
\textbf{TAR-ViTPose*} & ViT-S     & 81.6          & 84.7          & 80.9          & 74.2          & 77.8          & 77.9          & 71.7          & 78.5          \\
\rowcolor[gray]{0.9}
\textbf{TAR-ViTPose*} & ViT-B     & 82.9          & 87.8          & 85.9          & 82.1          & 80.1          & 84.5          & 82.0          & 83.6          \\
\rowcolor[gray]{0.9}
\textbf{TAR-ViTPose*} & ViT-H     & \textbf{94.3} & \textbf{93.3} & \textbf{91.3} & \textbf{88.8} & \textbf{88.7} & \textbf{90.2} & \textbf{88.6} & \textbf{91.0} \\
\hline
\end{tabular}
\caption{Comparison with the SOTAs on PoseTrack21 val. set. ‘\#’ indicates that the ViT backbone used is not specified, `$\dagger$' indicates the end-to-end approach. Similar to FAMI-Pose \cite{FAMIPose_CVPR2022} and DSTA \cite{DSTA-CVPR2024}, our proposed TAR-ViTPose sets the temporal span \(T\) to 2, which includes two preceding and two succeeding frames, totaling four auxiliary frames.}
\label{tab:compare_to_sota_21}
\vspace{-2mm}
\end{table}

\section*{C. Additional Ablation Study}
\begin{table}[t]
\centering
\begin{tabular}{c|ccccc}
\hline
Threshold $\phi$ & 0.1  & 0.15 & 0.2         & 0.25 & 0.3  \\ \hline
mAP            & 83.2 & 83.7 & \textbf{84.0} & 83.5 & 82.9 \\ \hline
\end{tabular}
\caption{Impact of different mask thresholds.}
\label{tab:mask_threshold}
\end{table}

\begin{table}[t]
\centering
\begin{tabular}{c|ccccc}
\hline
\#Layer & 2    & 4    & 6  & 8    & 10   \\ \hline
mAP        & 83.4 & 83.8 & \textbf{84.0} & 83.9 & 83.9 \\ \hline
\end{tabular}
\caption{Different number of layers in JTA.}
\label{tab:jta_num_layer}
\end{table}

\begin{table}[t]
\centering
\begin{tabular}{c|cccc}
\hline
\#Layer  & 1    & 2    & 3    & 4    \\ \hline
mAP             & 84.0 & 83.9 & 84.0 & 84.0 \\ \hline
\end{tabular}
\caption{Different number of layers in GRA.}
\label{tab:gra_num_layer}
\end{table}

\textbf{Mask Threshold.} We further analyze the influence of the mask threshold $\phi$ in Eq.~\ref{eqn:mask_attention} on constructing the joint-specific binary masks. As shown in Table~\ref{tab:mask_threshold}, masking thresholds in the range of 0.1 to 0.25 all yield high human pose estimation accuracy. Intuitively, a lower threshold enlarges the attention region around each joint's location in successive frames, while a higher threshold produces a smaller and more concentrated focus area. With a threshold of {0.2}, the model achieves an optimal balance, reaching {84.0} mAP.

\vspace{2mm}
\noindent\textbf{Number of Layers in JTA.}
Table~\ref{tab:jta_num_layer} analyzes the effect of varying the number of layers in our $\text{JTA}(\cdot,\cdot)$ module. Increasing the depth from 2 to 6 layers consistently improves performance (83.4 → 84.0 mAP), which is consistent with the general observation that deeper architectures offer stronger representation capacity. However, further increasing the number of layers beyond 6 yields no additional benefit and even leads to a slight performance drop (83.9 mAP with 8 or 10 layers). We attribute this saturation to the redundancy introduced by excessive temporal aggregation, which may over-smooth or dilute frame-specific cues. Overall, a depth of 6 layers provides the best balance between accuracy and efficiency, and is therefore adopted as the default configuration in our main experiments.

\begin{figure*}
    \centering
    \includegraphics[width=1.0\linewidth]{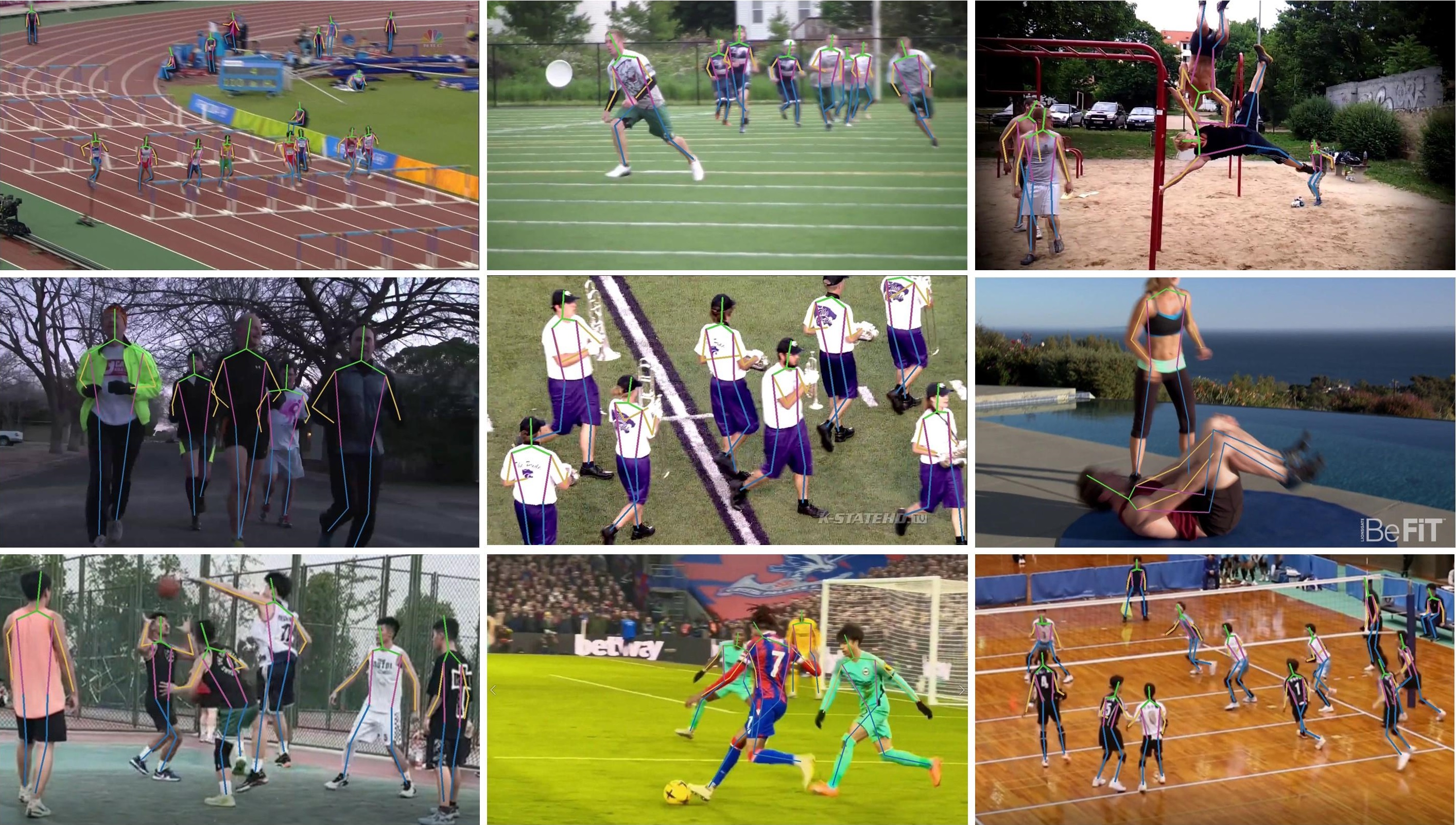}
    \caption{Additional qualitative results of our TAR-ViTPose. The first two rows are from the PoseTrack datasets, while the last row is from in-the-wild videos. Across all examples, the model remains robust under occlusion, motion blur, complex poses, and defocus.}
    \label{fig:appendix_qualitative}
    \vspace{-2mm}
\end{figure*}

\vspace{2mm}
\noindent\textbf{Number of Layers in GRA.}
In our GRA module, we use a single cross-attention layer to inject the aggregated temporal information back into the feature space of the current frame. As shown in Table~\ref{tab:gra_num_layer}, using only one layer already achieves strong performance (84.0 mAP), and increasing the number of layers does not provide any additional improvement. We attribute this to the nature of GRA: it does not extract or aggregate new features but simply restores the joint-specific temporal features to the current-frame representation, a task that can be effectively accomplished with a shallow design. Therefore, we adopt a single cross-attention layer as the default configuration to maintain high accuracy while achieving optimal efficiency.

\vspace{2mm}
\noindent\textbf{Auxiliary Frame.} In addition, we investigate the impact of using different numbers of auxiliary frames. The results reported in Table~\ref{tab:diff_aux_frame} consistently show that increasing the number of auxiliary frames leads to improved performance across ViT backbones of different scales. This observation aligns with our intuition that incorporating more auxiliary frames provides richer and more complementary temporal information, thereby enabling more accurate and robust pose estimation for the key frame.

\begin{table}[t]
\centering
\begin{tabular}{c|ccc}
\#Auxiliary Frame    & ViT-S             & ViT-B             & ViT-H \\ \hline
1 \{-1\}             & 81.4              & 83.2              & 86.1  \\
2 \{-1, +1\}         & 81.5              & 83.4              & 86.3  \\
4 \{-2, -1, +1, +2\} & \textbf{81.9}     & \textbf{84.0}     & \textbf{86.8} 
\end{tabular}
\caption{Different number of auxiliary frames. `-' indicates previous frames while `+' indicates subsequent frames.}
\label{tab:diff_aux_frame}
\end{table}

\begin{figure}
    \centering
    \includegraphics[width=1.0\linewidth]{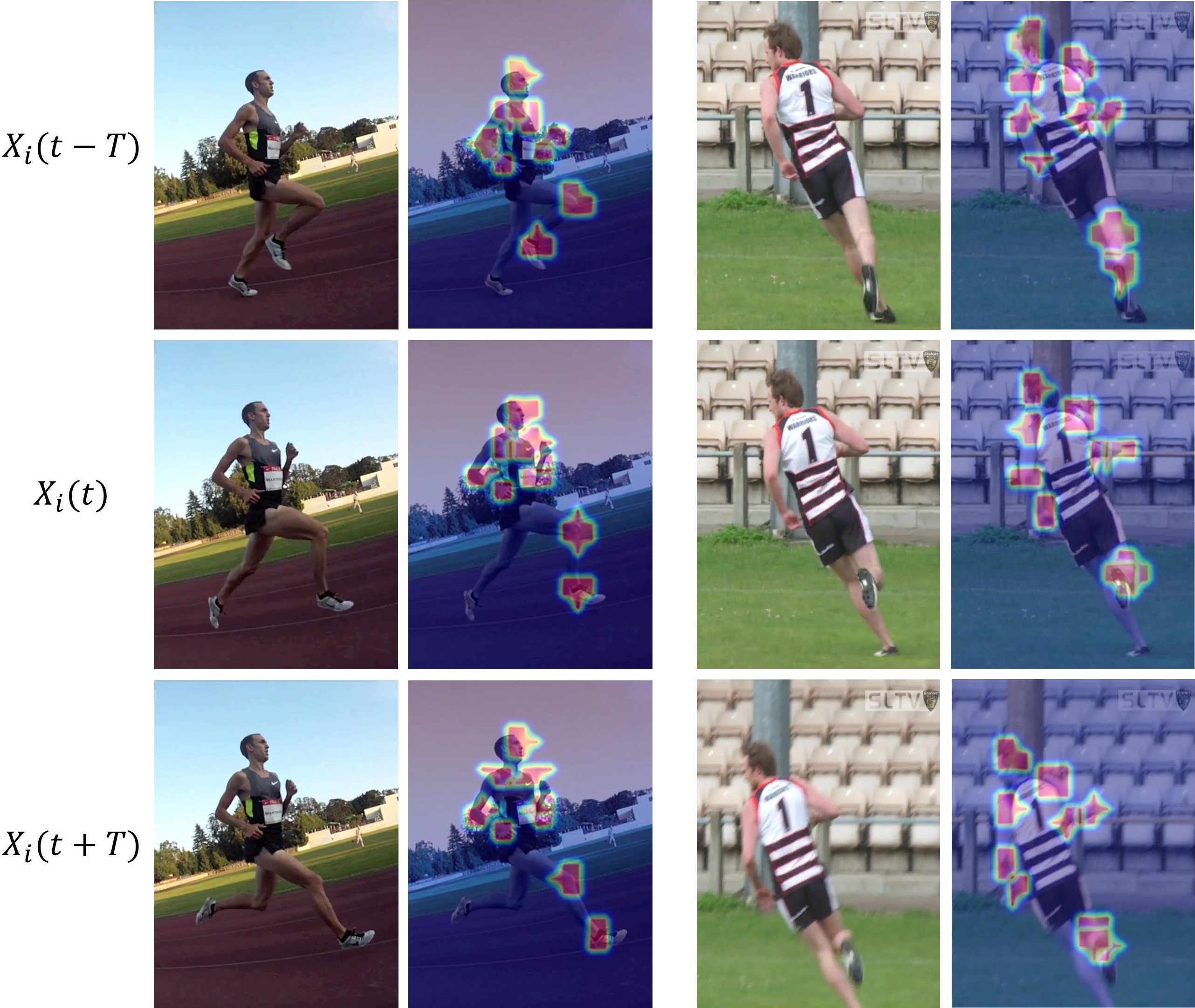}
    \caption{Visualization of attention heatmaps for joint query tokens with mask-aware attention. For each example, given a current frame \(X_i(t)\) and its neighboring frames \(X_i(t-T)\) and \(X_i(t+T)\), we visualize the attention heatmaps of nine joint query tokens, including the head, shoulders, elbows, wrists, right knee, and right ankle, over the feature maps of these three frames.}
    \label{fig:appendix_attention}
    \vspace{-2mm}
\end{figure}

\section*{D. Qualitative Results}
\textbf{D1.} Additional qualitative results on the PoseTrack validation sets and in-the-wild videos are presented in Fig.~\ref{fig:appendix_qualitative}. More results are provided in the accompanying video.

\vspace{2mm}
\noindent \textbf{D2.} To further validate that the proposed mask-aware attention enables each joint query token to effectively focus on its corresponding keypoint regions across frames while suppressing interference from unrelated areas, additional attention heatmaps of joint query tokens are provided in Fig.~\ref{fig:appendix_attention}.

\section*{E. Limitation and Future Work}

It is important to clarify that our work does not target temporal pose tracking. Instead, it introduces a simple yet robust temporal Vision Transformer framework that delivers strong performance for 2D pose estimation in videos. Because the method does not explicitly impose temporal consistency across frames, it may occasionally produce slight temporal inconsistencies, especially in heavily occluded scenes, as shown in the supplementary video.
Looking ahead, we plan to extend our framework to multi-person pose tracking by incorporating temporal identity consistency.